\documentclass[sigconf]{acmart}

\AtBeginDocument{%
  \providecommand\BibTeX{{%
    \normalfont B\kern-0.5em{\scshape i\kern-0.25em b}\kern-0.8em\TeX}}}

\copyrightyear{2023}
\acmYear{2023}
\setcopyright{rightsretained}
\acmConference[KDD '23]{Proceedings of the 29th ACM SIGKDD Conference on Knowledge Discovery and Data Mining}{August 6--10, 2023}{Long Beach, CA, USA}
\acmBooktitle{Proceedings of the 29th ACM SIGKDD Conference on Knowledge Discovery and Data Mining (KDD '23), August 6--10, 2023, Long Beach, CA, USA}
\acmDOI{10.1145/3580305.3599278}
\acmISBN{979-8-4007-0103-0/23/08}

\usepackage{etoolbox}
\makeatletter
\patchcmd{\maketitle}{\@copyrightpermission}{
   \begin{minipage}{0.3\columnwidth}
     \href{https://creativecommons.org/licenses/by/4.0/}{\includegraphics[width=0.90\textwidth]{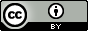}}
   \end{minipage}\hfill
   \begin{minipage}{0.7\columnwidth}
     \href{https://creativecommons.org/licenses/by/4.0/}{This work is licensed under a Creative Commons Attribution International 4.0 License.}
   \end{minipage}

   \vspace{5pt}
}{}{}

\makeatother

\usepackage{paralist}
\usepackage{subcaption}
\usepackage{multirow}
\usepackage{enumitem}
\usepackage{amsfonts,bm}

\def\eqref#1{equation~\ref{#1}}

\def\1{\bm{1}}

\usepackage{tablefootnote}

\DeclareMathAlphabet{\mathsfit}{\encodingdefault}{\sfdefault}{m}{sl}
\SetMathAlphabet{\mathsfit}{bold}{\encodingdefault}{\sfdefault}{bx}{n}

\newcommand{\R}{\mathbb{R}}

\DeclareMathOperator*{\argmax}{arg\,max}

\newcommand{\norm}[1]{\left\lVert#1\right\rVert}
\newcommand{\B}{\boldsymbol}

\newcommand{\MC}{\mathcal}
\usepackage{pifont}

\newcommand{\modelaname}{\textit{COMET}}

\newcommand{\modelbname}{\textit{COMET+}}

\newcommand{\modelcname}{\textit{Top-k+}}

\newcommand{\modeldname}{\textit{Hash-r+}}

\newcommand{\modelename}{\textit{COMET-BERT}}

\makeatletter  
\usepackage{stmaryrd}  
\newcommand{\fixed@sra}{$\vrule height 2\fontdimen22\textfont2 width 0pt\shortrightarrow$}
\newcommand{\shortarrow}[1]{%
  \mathrel{\text{\rotatebox[origin=c]{\numexpr#1*45}{\fixed@sra}}}
}
\makeatother

\usepackage{wrapfig}

\begin{document}

\title[\modelaname~and Permutation-Based Local Search]{COMET: Learning Cardinality Constrained Mixture of Experts with Trees and Local Search}


\author{Shibal Ibrahim}
\affiliation{%
  \institution{Massachusetts Institute of Technology}
  \city{Cambridge}
  \state{MA}
  \country{USA}
}
\email{shibal@mit.edu}

\author{Wenyu Chen}
\affiliation{%
  \institution{Massachusetts Institute of Technology}
  \city{Cambridge}
  \state{MA}
  \country{USA}
}
\email{wenyu@mit.edu}

\author{Hussein Hazimeh}
\affiliation{%
  \institution{Google Research}
  \city{New York}
  \state{NY}
  \country{USA}
}
\email{hazimeh@google.com}

\author{Natalia Ponomareva}
\affiliation{%
 \institution{Google Research}
 \city{New York}
 \state{NY}
 \country{USA}
 }
\email{nponomareva@google.com}

\author{Zhe Zhao}
\affiliation{%
  \institution{Google DeepMind}
  \city{Mountain View}
  \state{CA}
  \country{USA}
}
\email{zhezhao@google.com}

\author{Rahul Mazumder}
\affiliation{%
  \institution{Massachusetts Institute of Technology}
  \city{Cambridge}
  \state{MA}
  \country{USA}
}
\email{rahulmaz@mit.edu}

\renewcommand{\shortauthors}{Ibrahim et al.}

\begin{abstract}
The sparse Mixture-of-Experts (Sparse-MoE) framework efficiently scales up model capacity  in various domains, such as natural language processing and vision. Sparse-MoEs select a subset of the ``experts'' (thus, only a portion of the overall network) for each input sample using a sparse, trainable gate. Existing sparse gates are prone to convergence and performance issues when training with first-order optimization methods. In this paper, we introduce two improvements to current MoE approaches. First, we propose a new sparse gate: \modelaname, which relies on a novel tree-based mechanism. \modelaname~is differentiable, can exploit sparsity to speed up computation, and outperforms state-of-the-art gates. Second, due to the challenging combinatorial nature of sparse expert selection, first-order methods are typically prone to low-quality solutions. To deal with this challenge, we propose a novel, permutation-based local search method that can complement first-order methods in training  \textit{any} sparse gate, e.g., Hash routing, Top-k, DSelect-k, and \modelaname. We show that local search can help networks escape bad initializations or solutions. We performed large-scale experiments on various domains, including recommender systems, vision, and natural language processing. On standard vision and recommender systems benchmarks, \modelbname~(\modelaname~with local search) achieves up to 13\% improvement in ROC AUC over popular gates, e.g., Hash routing and Top-k, and up to 9\% over prior differentiable gates e.g., DSelect-k. When Top-k and Hash gates are combined with local search, we see up to $100\times$ reduction in the budget needed for hyperparameter tuning. Moreover, for language modeling, our approach improves over the state-of-the-art MoEBERT model for distilling BERT on 5/7 GLUE benchmarks as well as SQuAD dataset.
\end{abstract}





\maketitle

\section{Introduction}
The Sparse Mixture of Experts (Sparse-MoE) framework has led to state-of-the-art performance in various applications such as natural language processing (NLP) \citep{Shazeer2017,Fedus2021,Zoph2022,Du2022,Artetxe2022}, vision \citep{Ruiz2021,Wu2022}, time-series analysis \citep{Ismail2022}, multi-task learning \citep{ma2018modeling,Kudugunta2021,Hazimeh2021}, and multimodal learning \citep{Mustafa2022}. 
Sparse-MoE consists of a set of trainable experts (neural networks) and a trainable sparse gate.
The sparse gate in Sparse-MoE selects an appropriate subset of experts on a per-sample basis, which allows for faster computation \citep{Shazeer2017} and enhances interpretability \citep{Fedus2021,Ismail2022}.

The literature on Sparse-MoE has traditionally focused on Top-k gating, which selects $k$ out of $n$ experts using a Top-k operation \citep{Shazeer2017,Fedus2021,Zoph2022}.
Top-k gating is simple and efficient because it allows sparse training.
However, as highlighted by prior literature \citep{Fedus2021,Zoph2022,Hazimeh2021}, the non-continuous nature of Top-k makes it susceptible to stability and convergence issues.
Alternative gating strategies exist in the literature, based on
reinforcement learning \citep{Bengio2016} or post-processing via linear assignment \citep{Lewis2021,Clark2022}.
However, these strategies also face challenges in terms of efficiency and 
interpretability; see related work in Section \ref{sec:related-work} for more details.
Random routing strategies \citep{Roller2021,Zuo2022} alternatively bypass learning of the gating function altogether. 
Although computationally efficient, these strategies lead to performance degradation \citep{Clark2022}. 
Recent work
\citep{Hazimeh2021} demonstrates that differentiable gating in Sparse-MoE can  improve stability and  performance compared to popular non-differentiable gates. 
However, it suffers from expert collapse in some cases as we observed in our experiments.
 
In this paper, we propose two new approaches for improving routing in Sparse-MoE. 
First, we introduce a novel differentiable sparse gate \modelaname\footnote{\modelaname: This stands for \underline{C}ardinality 
c\underline{O}nstrained 
\underline{M}ixture of \underline{E}xperts with \underline{T}rees.} that improves over 
existing state-of-the-art sparse gates \citep{Roller2021,Shazeer2017,Fedus2021,Zoph2022,Hazimeh2021}.
Second, we argue that the combinatorial nature of expert selection in Sparse-MoE presents a serious challenge for first-order methods. In particular, the performance of these methods is highly dependent on initialization, and they can get stuck in low-quality routing solutions.
Thus, we propose a new permutation-based local search method for Sparse-MoEs, which can help first-order methods  escape low-quality initializations or solutions. Our local search approach is general and can be applied to any sparse gate, including Top-k \citep{Shazeer2017}, Hash routing \citep{Roller2021}, DSelect-k \citep{Hazimeh2021}, and our proposed gate \modelaname.

\paragraph{\modelaname} Our proposed \modelaname~gate is the first decision-tree-based selection mechanism for sparse expert selection --- decision trees naturally perform per-sample routing (i.e., each sample follows a root-to-leaf path). 
Our gate has several advantages:
(i) it is differentiable and can be optimized using first-order optimization methods e.g., stochastic gradient descent;
(ii) it allows (partially) conditional training, i.e., dense-to-sparse training;
(iii) it enforces a cardinality constraint, i.e., selects (at most) k out of the n experts;
(iv) it has superior predictive performance over state-of-the-art gates such as Hash routing, Top-k, and DSelect-k.

\paragraph{Local Search}
The learning problem underlying Sparse-MoEs is of combinatorial nature, which poses additional challenges compared to non-MoE machine learning models.
Popularly used optimization methods, such as SGD, may lead to  low-quality  solutions in Sparse-MoE, as we demonstrate in our numerical experiments in Section \ref{sec:experiments}.
To this end, we propose a permutation-based local search method, which can help first-order methods escape bad initializations and lead to better sample routing for any sparse gate e.g., Top-k, Hash routing, DSelect-k and even  \modelaname.
To the best of our knowledge, we are the first to explore local search methods in the context of Sparse-MoE.  
We provide empirical evidence through ablation studies and large-scale experiments to demonstrate permutation-based local search (i) pushes learning towards better gate/expert initializations in early optimization stages (see Section~\ref{sec:ablation-studies});
(ii) effectively reduces the budget needed for hyperparameter tuning by up to $100\times$ for some popular gates e.g., Hash Routing and Top-k (see Section~\ref{sec:bootstrap-trials});
(iii) leads to SOTA performance in terms of prediction and expert selection when combined with \modelaname, across various applications (see Section~\ref{sec:experiments}).

\paragraph{\textbf{Contributions}}
As discussed earlier,  it is well-known in the literature that popular sparse gates are challenging to train and may suffer from stability and performance issues. In this context, our contributions can be summarized as follows:
\begin{itemize}[noitemsep,topsep=0pt,parsep=0pt,partopsep=0pt, leftmargin=*]
\item 
We propose \modelaname, a novel tree-based sparse gate  
that simultaneously has the following desirable properties:
(a) differentiable,
(b) allows (partially) conditional training i.e., dense-to-sparse training, and sparse inference,
(c) satisfies per-sample cardinality constraint (selects at most $k$ out of the $n$ experts per-sample, where $k$ is a user-specified parameter).
\item Popular first-order methods used to optimize Sparse-MoEs are heavily influenced by expert and gate initializations, and may get stuck in low-quality solutions. 
Hence, we introduce a novel permutation-based local search method that can complement first-order methods by helping them escape bad initializations or solutions. Our local search method is general and can be applied to any gate, e.g., Hash routing, Top-k, and \modelaname.
\item We perform extensive experiments on recommender systems, vision and natural language processing tasks to highlight that \modelaname~ and \modelbname~(\modelaname~combined with local search) can give boosts in predictive performance.
In particular, on recommender and image datasets, we observed that \modelbname~can improve AUC performance by up to 13\% over existing sparse gates e.g., Top-k and Hash routing.
It can also reduce tuning by up to $100\times$ over popular gates e.g., Hash routing, and Top-k.
Similarly, in natural language processing applications, our \modelename~model (MoE based variant of BERT with \modelaname / \modelbname~ gating) can outperform state-of-the-art Hash-routing-based MoEBERT model \citep{Zuo2022moebert} on 5/7 GLUE benchmarks as well as SQuAD dataset for distilling pre-trained BERT model \citep{Devlin2018}.
\end{itemize}

\section{Related work}
\label{sec:related-work}
\paragraph{Sparse-Mixture-of-Experts}~~
The MoE framework was introduced by \cite{Jacobs1991}, and since then has been extensively studied --- see e.g., \cite{Jordan1993,Jacobs1997,Jiang1999}.
More recently, \cite{Shazeer2017} proposed a \textit{Sparse}-MoE framework,  based on the Top-k gate, and showed good performance on natural language processing tasks.
It was further improved upon by \cite{Fedus2021,Zoph2022,Zhou2022}.
However, Top-k gate
does not optimize the core expert selection problem as pointed out by \cite{Clark2022}.
Additionally, as highlighted by prior literature \cite{Fedus2021,Zoph2022,Hazimeh2021}, the non-continuous nature of Top-k makes it vulnerable to training stability and convergence issues.

With \emph{BASE} Layers, \cite{Lewis2021,Clark2022} formulate Sparse-MoE as an assignment problem where they post-process the gate output for balanced expert selection. 
\cite{Bengio2016} formulates the expert selection as a reinforcement learning problem. Others \cite{Roller2021,Zuo2022} proposed random routing strategies that do not learn the gating function during training. These methods are also promising as they have been shown to outperform models that learn routing through Top-k, e.g., in Switch Transformers \cite{Fedus2021}. 
Lastly, \cite{Hazimeh2021} introduced DSelect-k, a differentiable gate based on binary encodings, which improves over Top-k in terms of stability and statistical performance.

\paragraph{Conditional Computation}~~ In addition to the Sparse-MoE framework, there are other related works that also study conditional computation, i.e., the setup where only some parts  of neural network are activated based on the input --- see e.g., \cite{Bengio2013,Bengio2016,Ioannou2016,Wang2018}. These works rely on heuristics where the training and inference models are different.
More recently, \cite{Hazimeh2020} introduced conditional computation in differentiable (a.k.a. soft) trees \cite{Jacobs1991,Irsoy2012,Frosst2017,Hehn2019,Hazimeh2020,Ibrahim2022}.
Their proposal
allows routing samples through small parts of the tree; thus allowing for conditional computation with customized algorithms.
Our work builds upon this approach to solve the cardinality-constrained expert selection problem in Sparse-MoE.
Note that \cite{Hazimeh2020} does not address sparse expert selection in Sparse-MoE.

\paragraph{Local Search and Permutation Learning}~~
There is an extensive optimization literature on local search, e.g.,  \cite{Beck2013,HazimehL0Learn}. However, such methods have not been used in Sparse-MoE. Here, we survey permutation learning methods that are most relevant to our proposal.
This work uses differentiable relaxations of permutation via Sinkhorn operators \cite{Adams2011,Mena2018}.
These earlier works use these relaxations in other contexts e.g., ranking in \cite{Adams2011} and sorting in \cite{Mena2018}.
We use permutation learning as a local search to complement first-order optimization methods to improve sample routing in Sparse-MoE.

\section{Learning Sparse Mixture of Experts with decision trees}
\label{sec:sparse-moe}
\paragraph{Problem Setup of Sparse-MoE}
We first review the Sparse-MoE objective.
We assume that the task has an input space $\MC{X} \subseteq \R^p$ and an output space $\MC{Y} \subseteq \R^u$. Denote the $n$-dimensional simplex by $\Delta_n=\{w\in \R^n: \sum_{i \in [n]}w_i=1, w\geq 0\}$.  
In the MoE framework, the prediction function has two components: (i) a set of $n$ experts (parametrized by neural networks) $f_i:\MC{X}\to\R^u$ for any $i\in[n]:=\{1,2,\ldots,n\}$, and (ii) a gate $g:\MC{X}\to\Delta_n$ that outputs weights in the probability simplex.  Given a sample $x\in\MC{X}$, the corresponding output of the MoE is a convex combination of the experts with weights $g(x)$:   $\sum_{i \in [n]} f_i(x)g(x)_i.$ 

The goal of Sparse-MoE paradigm is to develop a gate that selects a convex combination of at most $k$ out of the $n$ experts.
The output of the gate can be thought of as a probability vector $g$ with at most $k$ nonzero entries, where $g(\cdot)_i$ is the weight assigned to the expert $f_i$.
The underlying optimization problem (also in \citep{Hazimeh2021}) is:
\begin{subequations}
\label{eq:sparse-moe}
\begin{align}
    \min_{f_1,\cdots,f_n, g} &\frac{1}{N} \textstyle\sum\limits_{(x,y)\in\MC{D}} \ell\left(y, \textstyle\sum\limits_{i \in [n]} f_i(x)g(x)_i\right), \\
    \text{s.t.}~~~~& \norm{g(x)}_0 \leq k,~~ g(x) \in \Delta_n, ~~\forall x \in \MC{X}. \label{eq:sparse-moe-sparse-simplex}
\end{align}
\end{subequations}
$\norm{g(\cdot)}_0$ denotes the number of nonzero entries in the vector $g(\cdot)$, $\ell(\cdot,\cdot)$ is the associated loss function such that $\ell: \MC{Y} \times \MC{X} \rightarrow \R$, and $N$ denotes the size of training samples $\MC{D} = \{(x_i, y_i) \in \MC{X} \times \MC{Y}\}_{i=1}^N$.  

The cardinality constraint in (\ref{eq:sparse-moe-sparse-simplex}) ensures that the gate selects at most $k$ experts.
Some popular gates e.g., Top-k impose exact cardinality constraint instead of an inequality constraint in (\ref{eq:sparse-moe-sparse-simplex}). 
However, the inequality constraint can allow for sparser expert selection as observed in prior work \citep{Hazimeh2021} and in our experiments (Section \ref{sec:experiments}). 
Problem (\ref{eq:sparse-moe}) is a combinatorial optimization problem that is not amenable to stochastic gradient descent due to the cardinality constraint in (\ref{eq:sparse-moe-sparse-simplex}).
In the next sections, we discuss our formulation that ensures the cardinality constraint and the simplex constraints are satisfied despite optimization with gradient-based methods.

The rest of this section is organized as follows.
In Section \ref{sec:sparse-moe-with-trees}, we discuss a high-level overview of our novel tree-based framework, which equivalently sets up the cardinality-constrained objective in problem (\ref{eq:sparse-moe}) as a weighted sum of decision trees.
Next in Section \ref{sec:differentiable-trees-with-conditional-computing}, we provide background on a single decision tree that selects a single expert per-sample while (i) allowing for smooth  optimization, and (ii) conditional computation support --- routing samples to a single leaf.
Later in Section \ref{sec:cardinality-with-k-trees}, we dive deeper into our novel tree-based framework that combines such trees to satisfy the cardinality constraint for $k\geq1$ without violating the simplex constraint.
We additionally highlight important aspects regarding leaf parameterization and regularization.
Next, in Section \ref{sec:non-powers-of-2}, we discuss how our method handles settings where experts are non-powers of 2. 
We then discuss in Section \ref{sec:stable-numerical-implementation} an implementation of \modelaname~for numerically stable training.

\subsection{Sparse-MoE with $k$ decision trees}
\label{sec:sparse-moe-with-trees}
The cardinality constrained MoE objective (\ref{eq:sparse-moe}) can be formulated equivalently using a set of decision trees. 
Classical decision trees are naturally suited to route each sample to a single leaf with a chain of hierarchical decisions.
In the case of $k=1$, we propose a single decision tree to route samples, where each leaf node is associated with an expert.
In cases where $k>1$, we instantiate $k$ different decision trees and combine their output in a way that enforces the cardinality and simplex constraints in (\ref{eq:sparse-moe-sparse-simplex}).

Given that classical decision trees are not amenable to differentiable training with first-order methods, we use a variant \citep{Hazimeh2020} of differentiable (a.k.a. soft) decision trees \citep{Jacobs1991,Kontschieder2015,Hehn2019,Hazimeh2020,Ibrahim2022}.
We build upon this work to solve the cardinality-constrained problem (\ref{eq:sparse-moe}). 
We first provide a summary of a single soft tree (with conditional computation support) in Section \ref{sec:differentiable-trees-with-conditional-computing}.
This serves as a building block for selecting a single expert per-sample.

\begin{figure}[!t]
\centering
\includegraphics[width=\columnwidth]{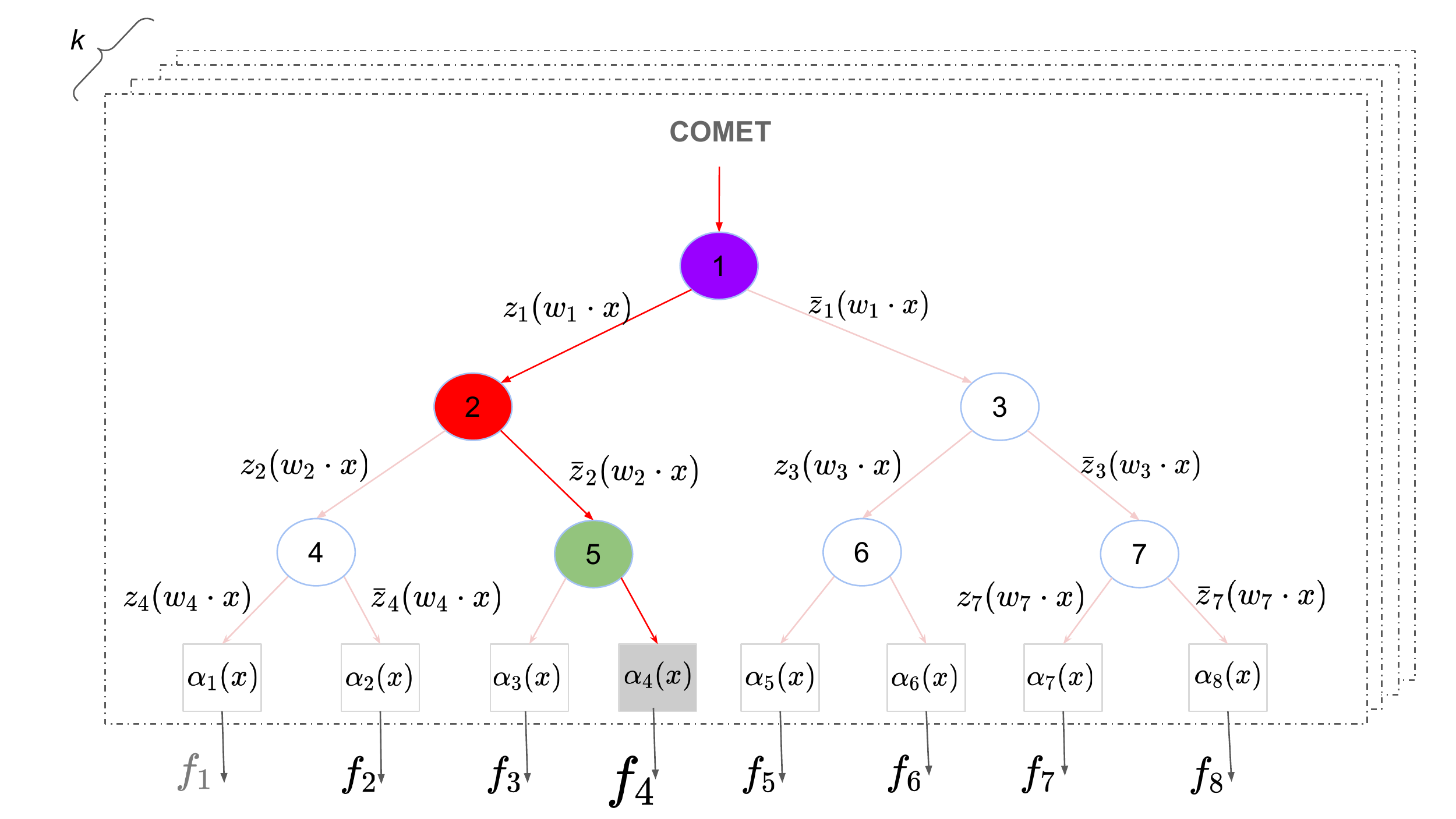} 
\caption{\modelaname~for 8 experts. 
Note $z_q(w_q \cdot x)$ denotes the binary state $\{0,1\}$ for $h(w_q \cdot x)$ achieved due to smooth-step activation function and entropic regularization.} 
\label{fig:comet-8}
\end{figure}

\subsection{Preliminaries: Differential Decision Tree with Conditional Computation}
\label{sec:differentiable-trees-with-conditional-computing}
In this section, we provide a brief summary of a variant \citep{Hazimeh2020} of a differentiable (a.k.a. soft) tree \citep{Jordan1993,Jacobs1991,Irsoy2012,Kontschieder2015,Hehn2019}, which we use to enable single-expert selection in Sparse-MoEs.
We extend it in the next section to solve the cardinality constrained problem for a general case $k\geq1$. 

\looseness=-1 Differentiable decision trees are similar to classical decision trees with hyperplane splits \citep{Murthy1994}.
However, they route each sample to left and right with different proportions, i.e., each sample reaches all leaves.   
Traditionally, differentiable decision trees have been unamenable to conditional computation as they cannot route a sample exclusively to the left or to the right.
Recent work \citep{Hazimeh2020} introduced a variant of the differentiable tree model that supports conditional computation.
Here, we discuss a brief summary of this variant.

We denote a single tree by $v:\MC{X}\to\Delta_n$, which maps an input sample $x\in\MC{X}$ to a probability vector $v$ over $\Delta_n$. Here, $n$ corresponds to the number of root-to-leaf paths (also equal to number of experts in the MoE paradigm).
Let $v$ be a binary tree with depth $d$ --- note our framework can naturally support cases where number of experts is non-powers of $2$, see Section \ref{sec:non-powers-of-2} for more details.
Let $\MC{I}$ and $\MC{L}$ denote sets of the internal (split) nodes and the leaves of the tree, respectively.
For any node $q \in \MC{I} \cup \MC{L}$, we define $T(q)$ as its set of ancestors.
Let $\{x \rightarrow q\}$ denote that a sample $x \in \R^p$ reaches $q$.

\paragraph{Sample Routing} Following prior work \citep{Kontschieder2015,Hehn2019,Hazimeh2020}, we will discuss sample routing using a probabilistic model. 
While sample routing is discussed using probability, differentiable trees are deterministic.
Differentiable trees are based on hyperplane splits \citep{Murthy1994}, where a \emph{linear} combination of the features is used in making routing decisions.
In particular, we assign a trainable weight vector $w_q \in \R^p$ with each internal node, which parameterizes the node's hyperplane split.
Let $h: \R \rightarrow [0, 1]$ be an activation function.
Given a sample $x \in \R^p$, the probability that internal node $q$ routes $x$ to the left is defined by $h(w_q \cdot x)$.

Now we summarize how to model the probability that $x$ reaches a certain leaf $l$ \citep{Kontschieder2015,Hehn2019,Hazimeh2020}.
Let $[l \shortarrow{5} q]$ (resp. $[q \shortarrow{7} l]$) denote the event that leaf $l$ belongs to the left (resp. right) subtree of node $q \in \MC{I}$.
The probability that $x$ reaches $l$ is given by: $\Pr(\{x \rightarrow l\}) = \prod_{q \in T(l)} r_{q,l}(x)$, where $r_{q,l}(x)$ is the probability of node $q$ routing $x$ towards the subtree containing leaf $l$, i.e.,
$r_{q,l}(x) := h(w_q \cdot x)^{\mathbf{1}[l \shortarrow{5}q]}(1 - h(w_q \cdot x))^{\mathbf{1}[q \shortarrow{7}l]}$.
Note that the vector $v(x)$ given by
\begin{align}
\label{eq:definition-of-v}
v(x) = [\Pr(\{x \rightarrow l_1\}),\cdots,\Pr(\{x \rightarrow l_n\})]\in\Delta_n,
\end{align}
defines a per-sample probability distribution over the $n$ leaves (or experts). 

Next, we discuss how the split probabilities $\{h(w_q \cdot x),  1-h(w_q \cdot x)\}$ can achieve binary state with a particular choice of activation function $h$ --- this is crucial for achieving sparse expert selection (and conditional computation) in the Sparse-MoE paradigm.

\paragraph{Smooth-Step Activation Function}
The common choice for activation function $h$ in soft tree literature  is a logistic function\citep{Jordan1993, Kontschieder2015,Frosst2017,Hehn2019}. However, it can not perform hard routing i.e., output exact zeros. 
This implies that any sample $x$ will reach every node in the tree with a positive probability, leading to a dense $v$. 
\citep{Hazimeh2020} proposed a smooth-step activation function for a variant of soft trees --- see Appendix \ref{supp-sec:smooth-step} for details.
Despite being continuously differentiable, smooth-step activation function can produce a sparse $v$ (after an initial warm-up period of soft routing) for hard routing. 
This is crucial for a sparse expert selection in Sparse-MoE paradigm.
Additionally, this choice of activation function also allows for (partially) conditional training with customized sparse backpropagation algorithms in soft trees (as shown in \citep{Hazimeh2020}), which is an important consideration for training large-scale Sparse-MoE models.

For cardinality-constrained Sparse-MoE learning with trees (not studied in \citep{Hazimeh2020}), the goal for each tree is to perform hard routing for all samples.
Therefore, we add additional regularization on $\{h(w_q \cdot x),  1-h(w_q \cdot x)\}$ to encourage convergence of $v$ to a one-hot state (discussed in more detail in Section \ref{sec:cardinality-with-k-trees}).

\subsection{Cardinality constraint with $k$ trees}
\label{sec:cardinality-with-k-trees}
Next, we discuss how to achieve the cardinality constraint ($k\geq1$) in Sparse-MoE with decision trees in the presence of simplex constraint. 
This key ideas are given as follows:
\begin{itemize}[noitemsep,topsep=0pt,parsep=0pt,partopsep=0pt, leftmargin=*]
\item We consider $k$ decision trees, where each tree $j$ selects a single expert via 
$v^{(j)}(\cdot)$ as defined in (\ref{eq:definition-of-v}).
\item With the experts selected as above, we need to decide the relative weights assigned to each expert. This is done  
through auxiliary functions $\alpha^{(j)}(\cdot)$, where $\alpha_i^{(j)}$ is a linear function $\beta_{i}^{(j)}\cdot x$ of the input. $\alpha_i^{(j)}$ reflects a linear weighting function (in the log space) for $i$-th expert (or leaf) in $j$-th tree.
\end{itemize}
See Figure \ref{fig:comet-8} as an example. 
Next, we define the prediction function for Sparse-MoE with $k$ decision trees to form \modelaname.

\paragraph{\modelaname~Prediction with $k$ Out of $n$ Experts} 
The prediction function for Sparse-MoE with $k\geq1$ is a weighted sum of the predictions of $i$-th expert (or leaf) across $k$ trees.  To this end, we define the weight for $i$-th expert as follows
\begin{equation}\label{eq:COMET-weights}
    g(x;\alpha, v)_i=\frac{\textstyle\sum\limits_{j \in [k]}\exp(\alpha_{i}^{(j)}(x))v_i^{(j)}(x)}{\textstyle\sum\limits_{j\in [k]}\textstyle\sum\limits_{i \in [n]}\exp(\alpha_{i}^{(j)}(x))v_i^{(j)}(x)},
\end{equation}
where $v_i^{(j)}(x)$ is the probability that a sample $x$ will reach expert $f_i$ in the $j$-th tree. 
Using (\ref{eq:COMET-weights}), the prediction function for Sparse-MoE with $k\geq1$ is given by
$    \hat{y}=\sum_{i\in[n]}f_i(x)g(x;\alpha, v)_i.$

We present the following proposition (proof in Appendix \ref{supp-sec:proof-for-proposition}):
\begin{proposition}
For any $\alpha$, if $v^{(j)}$ outputs a binary vector for every $j$,
the function $g(x;\alpha,v)$ satisfies the cardinality and simplex constraints in (\ref{eq:sparse-moe-sparse-simplex}).
\label{prop:sparse-simplex}
\end{proposition}

\paragraph{Accelerating Convergence of $v^{(j)}$ to One-Hot Encoding with Entropic Regularization}
In the Sparse-MoE setup, the goal is to achieve a one-hot vector state for $v^{(j)}$ quickly --- this ensures the cardinality constraint (i.e., to select at most $k$ experts) is respected by the $k$ trees.
To encourage faster convergence towards a one-hot vector, we add a per-tree entropy regularizer, $\lambda \Omega(v^{(j)}(x))$ to the loss objective, where $\Omega(v^{(j)}(x)) = -\sum_{i \in [n]} v_i^{(j)}(x) \log (v_i^{(j)}(x))$.
Entropy regularizers are used in \cite{Mena2018,Hazimeh2021} to get binary representations .

\paragraph{Dense-to-Sparse Learning} 
\modelaname~supports conditional training only partially. 
At the start of training, it uses all the available experts as $v^{(j)}$ is completely dense, so conditional training is not possible.
As training proceeds, $v^{(j)}$ becomes sparser due to smooth-step activation function and entropic regularization, eventually achieving binary state.
From this stage onwards, the gate satisfies the cardinality constraint per-sample, i.e, each sample gets routed to at most $k$ experts.
Hence, sparse training can proceed to refine the solution quality.
Empirically, we observe that a small number of epochs are sufficient for the optimizer to reach the sparse training phase.

\subsection{Non-powers of 2} 
\label{sec:non-powers-of-2}
\looseness=-1 Typically, in Sparse-MoE, each expert is assigned to a separate machine for efficiency \citep{Fedus2021,Zoph2022}. 
This may mean that the number of experts could be defined by the number of machines --- machines may not necessarily be available in powers of 2.
Our gate naturally handles cases where the number of experts are not chosen to be powers of 2.
We propose merging child nodes at the leaf level. In such instances, we have  imperfect binary decision trees (Fig. \ref{fig:non-powers-of-2} in Appendix) with $n$ nodes, with $2^d-n$ nodes in the $(d-1)$-th level, and $2n-2^d$ nodes in the $d$-th level. Additional details are in Appendix \ref{supp-sec:non-powers-of-2}.
In contrast to other differentiable gates (e.g., DSelect-k \citep{Hazimeh2021}), our proposed gate \modelaname~does not require any additional regularization to encourage the simplex constraint in (\ref{eq:sparse-moe-sparse-simplex}).

\subsection{Stable numerical implementation}
\label{sec:stable-numerical-implementation}
Next, we discuss a stable numerical implementation of \modelaname~gate.
\modelaname~introduces additional exponential functions in the expert weights (or leaf nodes of the decision trees) --- see (\ref{eq:COMET-weights}). 
More exponential functions are known to cause instabilities in Sparse-MoE models. For example, \cite{Zoph2022} introduced router z-loss in Switch Transformers to encourage smaller logits.
However, this may have a performance tradeoff.
In our implementation of \modelaname, we can mitigate instability issues arising from additional exponential functions using the following approach: (i) convert root-to-leaf probabilities to the log-space, $\log v^{(j)}_i(x)$, (ii) compute $\alpha_{i}^{(j)}+\log v_i^{(j)}(x)$, (iii) subtract the maximum, i.e., $\max_{i,j} (\alpha_{i}^{(j)}+\log v_i^{(j)}(x))$ from each element, (iv) apply a two-way softmax operation to get $g(x)$.

\section{Local search}
\label{sec:permutation-based-local-search}
Expert selection is a challenging combinatorial problem  that is known to be NP-hard.  Although first-order heuristics can usually provide fast  solutions, they rely heavily on initialization and are sometimes prone to arriving at low-quality solutions.
To this end, we propose a permutation-based local search method that complements first-order methods in optimizing Sparse-MoEs.
In both large-scale experiments and ablation studies, we see that the incorporation of local search can improve the performance of \textit{any} gating method and can significantly reduce the number of tuning trials.

Our approach derives inspiration from the local search methods 
commonly used along with the first-order methods to help escape local minima in sparse linear models\citep{Beck2013,HazimehL0Learn}.
We note that this is the first attempt in the literature to incorporate local search methods in the context of Sparse-MoE. Moreover, unlike common local search methods in literature, our proposed  search method is differentiable.
We want to highlight that our local search method is useful for any existing sparse gate, e.g., Hash routing, Top-k, and our proposed \modelaname. We hypothesize that our permutation-based approach can help navigate the optimization loss surface for various gates.

The rest of the section is organized as follows. In section \ref{sec:refined-sparse-moe-with-permutation}, we formulate a refined cardinality-constrained Sparse-MoE objective with additional binary variables to add support for permutation-based local search. 
Then, in section \ref{sec:background-on-permutation}, we provide background on permutation and its differentiable relaxation.
Next in section \ref{sec:empirical-considerations}, we outline our differentiable optimization approach for the refined Sparse-MoE objective and some additional practical considerations for computational efficiency. 
Later, in Section \ref{sec:ablation-studies}, we provide an ablation study to support our hypothesis that the local search can help escape  bad initializations.

\subsection{Permutation-based Local Search}
\label{sec:refined-sparse-moe-with-permutation}
In this section, we formulate a refined objective for the cardinality-constrained Sparse-MoE objective that adds support for permutation-based local search.

Let us denote by $\mathcal{S}_n$ the set of all permutations of the set $[n]$. 
Given any permutation $\sigma\in\mathcal{S}_n$, we permute the $n$ experts accordingly and assign $i$-th weight $g(x)_i$ to $\sigma(i)$-th expert instead of $i$-th expert.
With this permutation, the prediction for Sparse-MoE could be written as:
$
\hat{y}=\sum_{i \in [n]} f_{\sigma(i)}(x)g(x)_i.
$
We note that due to symmetry between experts and weights, permuting the experts is essentially same as permuting the weights. To see this, we can write
$    \sum_{i \in [n]} f_{\sigma(i)}(x)g(x)_i = \sum_{j \in [n]} f_j(x)g(x)_{\sigma^{-1}(j)}$,
where $\sigma^{-1}$ is the inverse map of $\sigma$, which is also a permutation.

For a permutation $\sigma$, we can define a corresponding permutation matrix $\B P^{\sigma}$, by setting $P^{\sigma}[i,j]=\bm1\{\sigma(j)=i\}$, where $\bm1\{\cdot\}$ is an indicator function.  Then it is easy to see that 
$     \sum_{j \in [n]} f_j(x)g(x)_{\sigma^{-1}(j)} = \sum_{j \in [n]}f_j(x)(\B P^{\sigma}g(x))_j$.
The refined Sparse-MoE problem is 
\begin{subequations}
\label{eq:sparse-moe-perm-mat}
\begin{align}
    \min_{f_1,\cdots,f_n, g, \B P} \quad &\frac{1}{N} \textstyle\sum\limits_{(x,y)\in\MC{D}} \ell\left(y, \textstyle\sum\limits_{i \in [n]} f_i(x)(\B Pg(x))_i\right), \\
    \text{s.t.}&~~~~~~\|g(x)\|_0\leq k,~~g(x)\in\Delta_n,~~\forall x\in \MC{X}, \\
    &~~~~~~\B P\in\mathcal{P}_n^{\text{local}},\label{eq:sparse-moe-P-perm-mat}
\end{align}
\end{subequations} 
where $\mathcal{P}_n^{\text{local}}$ is a localized set of permutations in the full set of permutations, which we denote by $\mathcal{P}_n$.
For example, one may only allow for $\mathcal{P}_n^{\text{local}}=\mathcal{P}_2$ , which only allows interchanging (swapping) two columns similar to ``swap'' operations shown to be useful in the sparse regression literature \citep{HazimehL0Learn}. Besides optimizing the gates and experts, formulation (\ref{eq:sparse-moe-perm-mat}) performs local search by optimizing over the permutation matrix. Specifically, the goal of local search here is to find a permutation $P$ that leads to a better solution, i.e., one with a lower objective. Intuitively, if SGD is stuck at a low-quality solution, the permutation may be able to escape the solution by a better reordering of the experts.
Standard local search, e.g., bruteforce search may be computationally expensive. Therefore, we resort to a differentiable method that can be optimized efficiently.

\subsection{Preliminaries: Permutation and a differentiable relaxation} 
\label{sec:background-on-permutation}
In this section, we briefly summarize how the permutation learning problem is parameterized and later optimized.
To parametrize the permutation matrix in the problem, a natural consideration is through the linear assignment problem~\citep{Kuhn1955}. To illustrate this, consider $n$ people are to complete $n$ tasks and a matrix $\B{U} \in \R_{\geq0}^{n \times n}$, the goal is to assign each task to one person so as to maximize the utility given that the utility of assigning task $j$ to person $i$ is $U_{ij}$. This leads to the following optimization problem
\begin{align}
    M(\B{U}) = \argmax_{\B{P} \in \MC{P}_n}{\langle \B{P},\B{U}\rangle_F:=\textstyle\sum\limits_{i\in [n]}\textstyle\sum\limits_{j\in [n]}P_{ij}U_{ij}}.
    \label{eq:permutation-discrete}
\end{align}
The operator $M$ here is called the Matching operator, which maps a nonnegative matrix $\B U$ to a permutation matrix $\B P$. 

Problem (\ref{eq:permutation-discrete}) is a combinatorial optimization problem, which admits the following linear relaxation~\cite{bertsimas1997introduction}:
\begin{align}
    \max_{\B{B} \in \MC{B}_n}{\langle \B{P},\B{U}\rangle_F:=\textstyle\sum\limits_{i\in[n]}\textstyle\sum\limits_{j\in[n]}P_{ij}U_{ij}},
    \label{eq:permutation-LP}
\end{align}
where $\mathcal{B}_n$ denotes the set of double stochastic matrices $\mathcal{B}_n=\{\B B\in\mathbb{R}^{n\times n}:\sum_{i\in[n]}B_{ij}=1,\sum_{j\in[n]}B_{ij}=1,B_{ij}\in[0,1]\}$, which is a convex hull of the set of permutation matrices $\mathcal{P}_n$.

However, this is still not a differentiable parametrization as problem~(\ref{eq:permutation-LP}) might end up with multiple solutions. To this end, \citet{Mena2018} proposes a smooth version\footnote{Note that (\ref{eq:permutation-continuous}) is an entropy-regularized version of (\ref{eq:permutation-LP}). Since the entropy term is strictly concave, problem~(\ref{eq:permutation-continuous}) has a unique minimizer and thus the parametrization is differentiable.} of the permutation learning objective in (\ref{eq:permutation-LP}):
\begin{align}
    S(\B{U}/\tau) = \argmax_{\B{B} \in \MC{B}_n}{\langle \B{B},\B{U}\rangle_F} - \tau \textstyle\sum\limits_{i,j \in [n]} B_{ij}\log{B_{ij}},
    \label{eq:permutation-continuous}
\end{align} and solves it using Sinkhorn operator $S(\cdot)$~\citep{Adams2011}, defined by the following recursion:
\begin{subequations}
\label{eq:sinkhorn}
\begin{align}
    S^0(\B{U}) &= \exp(\B{U}), \\
    S^r(\B{U}) &= \MC{T}_{col}(\MC{T}_{row}(S^{r-1}(\B{U}) )), \label{eq:sinkhorn-normalization}\\
    S(\B{U}) &= \lim_{r\rightarrow\infty}S^r(\B{U}), \label{eq:sinkhorn-limit}
\end{align}
\end{subequations}
where $\MC{T}_{row}(\B{U}) = \B{U} \oslash (\B{U}\B{1}_{n} \B{1}_{n}^T)$, and $\MC{T}_{col}(\B{U}) = \B{U} \oslash (\B{1}_{n} \B{1}_{n}^T \B{U})$ are the row and column-wise normalization operators of a matrix, with $\oslash$ denoting the element-wise division and $\B{1}_n$ a column vector of ones. 
The sinkhorn procedure in (\ref{eq:sinkhorn}) allows differentiable training with first-order methods, making it appealing as a local search method for Sparse-MoE.

\looseness=-1 As shown in \citep{Mena2018}, $M(\B{U})$ can be obtained as $\lim_{\tau\to0^+}S(\B{U}/\tau)$, and thus $\lim_{\tau\to0^+,r\to\infty}S^r(\B{U}/\tau)$.
In practice, we set a max number of iterations $R$ for normalization in (\ref{eq:sinkhorn-normalization}) as well as a small positive number $\tau>0$, and use $S^R(\B{U}/\tau)$ to approximate the limit (\ref{eq:sinkhorn-limit}). In this way, we are able to parametrize  the permutation matrix $\B P$ in (\ref{eq:sparse-moe-perm-mat}) as a differentiable function $S^{R}(\B U/\tau)$ of learnable matrix $\B{U}$. 
However, additional considerations are needed
to ensure that a hard permutation matrix can be achieved quickly in a few epochs --- this is important in Sparse-MoE paradigm for computational reasons and a well-defined measure of sparsity. We discuss these in the next section.

\subsection{Practical considerations for optimization}
\label{sec:empirical-considerations}
Next, we discuss some empirical considerations for the end-to-end learning approach that are important for Sparse-MoE.

\paragraph{Need for a hard permutation matrix} We would like to have a hard permutation matrix at inference time and ideally during the course of training, for exact sparsity and computational efficiency considerations. 
First, the gate does not perform sparse inference if the learnt permutation matrix is not a hard matrix. For example, even if $g(\cdot)$ is sparse, the refined weights $\B P\cdot g(\cdot)$ are not a sparse vector if $\B P$ is not a binary matrix. This would result in a dense mixture of experts.
Second, some sparse gates perform dense-to sparse-training (partially conditional training), e.g., DSelect-k, \modelaname, or variants of Top-k \citep{Nie2021}.
If the learnt permutation matrix is not hard, then sparse training cannot proceed in the later stages of optimization.  
To this end, we employ a two-stage optimization approach: (i) in the first stage, we simultaneously train the network (experts and gates) and the permutation for a small number of  epochs.
(ii) In the second phase, the permutation matrix is fixed and only the remaining network (experts and gate) is trained. Therefore, local search is only used in the early stages of training.
Empirically, we observe that a small number of  epochs ($1-10$) is sufficient to learn a good permutation in the first stage and improve solution quality.
Since local search is restricted to the first stage, the computational efficiency of gates that perform dense-to-sparse training is not affected by much --- please refer to Appendex \ref{supp-sec:effect-of-local-search-on-computation} for additional discussion.

In the two-stage approach outlined above, 
there is a transition from a soft to a hard matrix between the two stages.
As we mentioned earlier, we use $S^R(\B{U}/\tau)$ to approximate $M(\B{U})$ as a limit of $R\to\infty,\tau\to 0^+$.
In practice, the transition could be not continuous, as this approximation does not always reach a hard permutation matrix given that $R$ is finite and $\tau$ is nonzero. 
Therefore, at the transition point, we propose to convert the ``soft'' permutation matrix $S^R(\B{U}/\tau)$ to a hard one via the linear assignment problem given in  (\ref{eq:permutation-discrete}), by invoking $\B{U}$ as $S^R(\B{U}/\tau)$.
In addition, empirically, small $R$ can lead to numerical instabilities for small $\tau$ \citep{Mena2018}. 
Therefore, to decrease deviance of $S^R(\B{U}/\tau)$ from the closest hard permutation matrix, we introduce two schedulers on $R$ and $\tau$ that increase $R$ for decreased $\tau$: (i) Ramp up (linearly) $R$ from $20$ to $150$, (ii) Ramp down (linearly in log-scale) $\tau$ from $10^{-3}$ to $10^{-7}$.     

\looseness=-1 Although the above schedulers decrease the deviance between soft and its closest hard permutation matrix at the transition point, the method still appears to suffer from pseudo-convergence.
In particular, we observed, some row-columns can converge to fractional entries i.e., a 2x2 sub-block having all entries with $0.5$.
Therefore, we introduce small separate row-wise and column-wise entropic regularizations to mitigate such degenerate cases: \\
$\zeta\sum_{i \in [n]}(\Omega(\mathcal{S}^R(\B{U}/\tau)_i) + \Omega(\mathcal{T}_{row}(\mathcal{S}^R(\B{U}/\tau))_i))$, where $\zeta \geq 0$.

\paragraph{Implicit localization} In the spirit of common local search approaches, a potential optimization approach could alternate between optimization of network (experts and gates) and permutation matrix.
However, this is unnecessary because the differentiable relaxation of permutation is also amenable to first-order methods. 
Therefore, our approach jointly optimizes both the network and the permutation matrix.     
We noted earlier that the search space for permutation is ``localized'' out of the full set of permutation matrices $\mathcal{P}_n$.
This localization is implicitly imposed through the smooth optimization of the permutation matrix via Sinkhorn.
The permutation matrix learning relies on the initialization for $\B{U}$ and at each gradient step the $\B{U}^{(t)}$ is naturally expected to not deviate drastically from $\B{U}^{(t-1)}$.
Since the permutation matrix is updated for a limited number of steps in first stage, intuitively it cannot deviate significantly from the initial permutation matrix. This also defines an implicit neighborhood.

\subsection{Ablation study for local search}
\label{sec:ablation-studies}
In this section, we provide an ablation study to provide evidence that the permutation-based local search can complement first-order optimization methods for routing in Sparse-MoE.
The study highlights that local search can improve solution quality through escape out of bad initializations in the first stages of optimization for 
different types of routing strategies: (a) fixed gates, (b) trainable gates.
We perform this study on a subsampled (200k) MovieLens dataset and use the same MoE architecture with 16 experts as the one described in Supplemental Section \ref{supp-sec:architectures}. 
We trained models for only 10 epochs without/with local search, where in the latter case we fixed the number of epochs for permutation learning to 5 epochs and $\zeta=10^{-5}$.
We used a batch size of $512$ and learning rate of $2.5\times10^{-5}$.
We repeat the training with 100 different random initializations and compute averages along with their standard errors.

\paragraph{Fixed Gates}
In fixed gating strategies e.g., random hash routing (Hash-r), the samples are pre-assigned to experts. For example, in natural language processing tasks, tokens or words in vocabulary are clustered randomly \citep{Roller2021} \emph{before} training begins into groups and each group of words are assigned to a random expert in the set of experts.
In our experiments on recommender systems, we randomly pre-assigned samples to experts based on user index for Hash-r (and \modeldname).
It is possible that the same group of users could be better aligned with another expert based on expert and user embedding initializations.
Permutation-based local search can potentially find better assignment of each group to a more suited expert.  
We provide empirical evidence to demonstrate that local search indeed can find better loss.
We report the average out-of-sample loss achieved by both Hash-r and \modeldname~in Table \ref{tab:initialization}.
Learning permutation appears to help map each pre-assigned cluster of users to a more suitable expert based on expert initialization for second stage of optimization.
\begin{table}[!t]
\small
\caption{Test loss ($\times10^{-2}$) achieved for different gates without and with (marked with $+$) local search in early stages of optimization.  Asterisk(*) indicates statistical significance (p-value<0.05) over the corresponding gate without permutation with a one-sided unpaired t-test.}
\label{tab:initialization}
\begin{tabular}{|l|l|l|c|}
\hline
Strategy & Smoothness & Gate & Test Loss $\downarrow$ \\ \hline
\multirow{2}{*}{Pre-assigned} & \multirow{2}{*}{-} & Hash-r & $~57.434\pm0.025$ \\
 &  & \modeldname & ${}^*\mathbf{57.000}\pm0.037$ \\ \hline
\multirow{4}{*}{Trainable} & \multirow{2}{*}{Non-differentiable} & Top-k & $~53.345\pm0.033$ \\
 &  & \modelcname & ${}^*\mathbf{53.140}\pm0.031$ \\ \cline{2-4} 
 & \multirow{2}{*}{Differentiable} & \modelaname & $~52.034\pm0.007$ \\
 &  & \modelbname & ${}^*\mathbf{52.017}\pm0.005$ \\ \hline
\end{tabular}
\end{table}

\paragraph{Trainable Gates}
For trainable gates, we also study the effect of local search on non-differentiable (Top-k) and differentiable gates (\modelaname).
We fixed $k=2$ for both types of gates and followed the same training protocol for 10 epochs. 
For \modelaname~(and \modelbname), we fixed $\gamma=0.01$ (for smooth-step) and $\lambda=1$ (for entropic regularization). For \modelcname and \modelbname, we fixed the number of epochs for permutation learning as 5.
We repeated this exercise for 100 different random initializations of the experts and gates. We report the average out-of-sample objective achieved by both types of gates in Table \ref{tab:initialization}.  
We can observe that local search appears to complement first-order optimization methods by learning better initializations in the first stage of Sparse-MoE optimization for later learning.

The practical significance of local search achieving a better test objective across many initializations for various gates can be seen in terms of reducing hyperparameter tuning overhead as discussed in Section \ref{sec:bootstrap-trials}.

\section{Experiments}
\label{sec:experiments}
We study the performance of \modelaname~and \modelbname~ in recommender systems and image datasets in Section \ref{sec:experiments-recommender-systems} and \modelename~in natural language processing tasks in \ref{sec:moebert}. We also study the effect of local search for various gates.
We denote our methods in \textit{italics}.
\subsection{Experiments on Recommender Systems and Image Datasets}
\label{sec:experiments-recommender-systems}
We study the performance of \modelaname~and \modelbname~in recommender systems and image datasets.
We compare with state-of-the-art gates and baselines including Softmax, Top-k, DSelect-k and Hash routing (Hash-r) on recommender systems (MovieLens \citep{movielens2016}, Jester\citep{Goldberg2001}, Books \citep{Ziegler2005}) and image datasets (Digits \citep{Deng2012,Netzer2011}, MultiMNIST \citep{Sabour2017}, MultiFashionMNIST \citep{Hazimeh2021}, CelebA\citep{Liu2015}).
We also include an ablation study in Section \ref{sec:reducing-tuning} that shows that \modelaname~achieves good performance with much less  trials than existing popular gates e.g., Hash routing and Top-k.
Additionally, in Section \ref{sec:bootstrap-trials}, we show that \modeldname,\modelcname, and \modelbname~with local search can potentially achieve good performance with much less trials than Hash-r, Top-k and \modelaname~respectively.

\paragraph{Implementation} We provide an open-source implementation of \modelaname~and \modelbname:  \url{https://github.com/mazumder-lab/COMET}.

\begin{table}[!b]
\centering
\caption{Tess Loss ($\times 10^{-2}$, the smaller the better) and number of experts per sample ($n/s$) for \modelaname, \modelbname~and benchmark gates across various recommender system datasets. Asterisk(*) indicates statistical significance (p-value<0.05) over the best existing gate, using a one-sided unpaired t-test.}
\label{tab:comet-rec-sys}
\setlength{\tabcolsep}{12pt}
\resizebox{\columnwidth}{!}{\begin{tabular}{|l|c|l|c|c|c|}
\hline
Dataset & $n$ & Model & Test Loss ~$\downarrow$ & $n/s$~$\downarrow$\\ \hline
\multirow{5}{*}{\begin{tabular}[c]{@{}l@{}}Books\\ ($\alpha=0.1$)\end{tabular}} & \multirow{5}{*}{9} & Softmax & $244.47\pm0.14$ & $9.00\pm0.00$ \\
 &  & Hash-r & $247.43\pm0.14$ & $~~1.00\pm0.00$ \\
 &  & \modeldname & $247.33\pm0.23$ & $~~1.00\pm0.00$ \\
 &  & Top-k & $247.87\pm0.17$ & $~~4.00\pm0.00$ \\
 &  & \modelcname & $247.88\pm0.14$ & $~~4.00\pm0.00$ \\
 &  & DSelect-k & $246.43\pm0.36$ & $~~1.09\pm0.00$ \\
 &  & \modelaname & ${}^*\textbf{240.79}\pm0.14$ & $~~2.81\pm0.11$ \\ 
&  & \modelbname & $240.82\pm0.19$ & $~~3.03\pm0.08$ \\ \hline
\multirow{5}{*}{\begin{tabular}[c]{@{}l@{}}Books\\ ($\alpha=0.9$)\end{tabular}} & \multirow{5}{*}{9} & Softmax & $~73.88\pm0.02$ & $9.00\pm0.00$ \\
 &  & Hash-r & $~75.02\pm0.03$ & $~~1.00\pm0.00$ \\
 &  & \modeldname & $~75.06\pm0.03$ & $~~1.00\pm0.00$ \\
 &  & Top-k & $~74.78\pm0.03$ & $~~4.00\pm0.00$ \\
 &  & \modelcname & $~74.86\pm0.03$ & $~~4.00\pm0.00$ \\
 &  & DSelect-k & $~~75.98\pm0.13$ & $~~1.07\pm0.00$ \\
 &  & \modelaname & $~~73.62\pm0.03$ & $~~2.94\pm0.10$ \\
&  & \modelbname & ${}^*\textbf{73.55}\pm0.02$ & $~~3.15\pm0.08$ \\ \hline
\multirow{5}{*}{\begin{tabular}[c]{@{}l@{}}MovieLens\\ ($\alpha=0.9$)\end{tabular}} & \multirow{5}{*}{16} & Softmax 
 & $~42.26\pm0.01$ & $16.00\pm0.00$ \\
 &  & Hash-r & $~46.91\pm0.02$ & $~~1.00\pm0.00$ \\
 &  & \modeldname & $~46.84\pm0.03$ & $~~1.00\pm0.00$ \\
 &  & Top-k & $~41.83\pm0.02$ & $~~2.00\pm0.00$ \\
 &  & \modelcname & $~41.74\pm0.02$ & $~~2.00\pm0.00$ \\
 &  & DSelect-k & $~40.82\pm0.02$ & $~~1.94\pm0.06$ \\
 &  & \modelaname & $~40.76\pm0.02$ & $~~1.76\pm0.06$ \\
&  & \modelbname & ${}^*\textbf{40.69}\pm0.02$ & $~1.66\pm0.06$ \\ \hline
\multirow{5}{*}{\begin{tabular}[c]{@{}l@{}}MovieLens\\ ($\alpha=0.1$)\end{tabular}} & \multirow{5}{*}{16} & Softmax & $~75.52\pm0.02$ & $16.00\pm0.00$ \\
 &  & Hash-r & $~79.41\pm0.02$ & $~~1.00\pm0.00$  \\
 &  & \modeldname & $~78.92\pm0.05$ & $~~1.00\pm0.00$  \\
 &  & Top-k & $~76.52\pm0.04$ & $~~2.00\pm0.00$ \\
 &  & \modelcname & $~75.12\pm0.04$ & $~~2.00\pm0.00$ \\
 &  & DSelect-k & $~73.91\pm0.05$ & $~~1.94\pm0.03$ \\
 &  & \modelaname & $~73.91\pm0.04$ & $~~1.94\pm0.03$ \\
 &  & \modelbname & ${}^*\textbf{73.67}\pm0.04$ & $~~1.98\pm0.03$\\ \hline
\multirow{5}{*}{\begin{tabular}[c]{@{}l@{}}Jester\\ ($\alpha=0.1$)\end{tabular}} & \multirow{5}{*}{16} & Softmax & $~68.17\pm0.03$ & $16.00\pm0.00$ \\
 &  & Hash-r & $~67.47\pm0.01$ & $~~1.00\pm0.00$ \\
 &  & Top-k & $~68.38\pm0.05$ & $~~2.00\pm0.00$ \\
 &  & \modelcname & $~68.00\pm0.07$ & $~~2.00\pm0.00$ \\
 &  & DSelect-k & $~67.06\pm0.03$ & $~~1.96\pm0.02$ \\
 &  & \modelaname & $~67.12\pm0.04$ & $~~1.98\pm0.02$ \\
 &  & \modelbname & ${}^*\textbf{66.91}\pm0.03$ & $~~2.00\pm0.00$ \\ \hline
 \multirow{5}{*}{\begin{tabular}[c]{@{}l@{}}Jester\\ ($\alpha=0.9$)\end{tabular}} & \multirow{5}{*}{16} & Softmax & $~21.936\pm0.002$ & $16.00\pm0.00$ \\
 &  & Hash-r & $~22.083\pm0.004$ & $~~1.00\pm0.00$ \\
 &  & Top-k & $~21.958\pm0.007$ & $~~2.00\pm0.00$ \\
 &  & \modelcname & $~21.961\pm0.006$ & $~~2.00\pm0.00$ \\
 &  & DSelect-k & $~21.930\pm0.005$ & $~~2.00\pm0.00$ \\
 &  & \modelaname & $~21.946\pm0.005$ & $~~2.00\pm0.00$ \\
 &  & \modelbname & ${}^*\textbf{21.906}\pm0.005$ & $~~2.00\pm0.00$ \\ \hline
 \end{tabular}}
\end{table}

\begin{table}[!b]
\centering
\caption{Tess Loss ($\times 10^{-2}$, the smaller the better) and number of experts per sample ($n/s$) for \modelaname, \modelbname~and benchmarks gates across various image datasets. Asterisk(*) indicates statistical significance (p-value<0.05) over the best existing gate, using a one-sided unpaired t-test.}
\label{tab:comet-image}
\setlength{\tabcolsep}{10pt}
\resizebox{\columnwidth}{!}{\begin{tabular}{|l|c|l|c|c|c|}
\hline
Dataset & $n$ & Model & Test Loss ~$\downarrow$ & $n/s$~$\downarrow$\\ \hline
\multirow{5}{*}{MultiFashionMNIST} & \multirow{5}{*}{5} & Softmax & $~~34.21\pm0.09$ & $~~5.00\pm0.00$ \\
 &  & Top-k & $~~33.82\pm0.09$ & $~~2.00\pm0.00$\\
 &  & \modelcname & ${}^{*}\textbf{33.62}\pm0.08$ & $~~2.00\pm0.00$\\
 &  & DSelect-k & $~~35.49\pm0.10$ & $~~1.00\pm0.00$\\
 &  & \modelaname & $~~33.70\pm0.09$ & $~~1.49\pm0.07$\\
 &  & \modelbname & ${}^{*}\textbf{33.67}\pm0.09$ & $~~1.54\pm0.07$\\ \hline
\multirow{5}{*}{CelebA} & \multirow{5}{*}{6} & Softmax & $~~35.10\pm0.32$ & $~~6.00\pm0.00$ \\
 &  & Top-k & $~~34.48\pm0.24$ & $~~2.00\pm0.00$\\
 &  & \modelcname & $~~34.54\pm0.23$ & $~~2.00\pm0.00$\\
 &  & DSelect-k & $~~35.39\pm0.12$ & $~~1.00\pm0.00$\\
 &  & \modelaname & $~~33.96\pm0.15$ & $~~1.00\pm0.08$\\
&  & \modelbname & ${}^{*}\textbf{33.93}\pm0.16$ & $~~1.00\pm0.08$\\ \hline
\multirow{5}{*}{Digits} & \multirow{5}{*}{8} & Softmax & $17.48\pm0.07$& $~~8.00\pm0.00$ \\
 &  & Top-k & $17.46\pm0.09$ & $~~2.00\pm0.00$\\
 &  & \modelcname & $17.29\pm0.08$ & $~~2.00\pm0.00$\\
 &  & DSelect-k & $17.18\pm0.06$ & $~~1.15\pm0.06$\\
 &  & \modelaname & $17.19\pm0.06$ & $~~1.07\pm0.04$\\
 &  & \modelbname & $\textbf{17.08}\pm0.09$ & $~~1.06\pm0.04$ \\ \hline
\multirow{5}{*}{MultiMNIST} & \multirow{5}{*}{16} & Softmax & $~~6.88\pm0.06$ & $~~16.00\pm0.00$ \\
 &  & Top-k & $~~6.84\pm0.05$ & $~~4.00\pm0.00$\\
 &  & \modelcname & $~~6.70\pm0.08$ & $~~4.00\pm0.00$\\
 &  & DSelect-k & $~~6.64\pm0.07$ & $~~3.40\pm0.09$\\
 &  & \modelaname & $~~\textbf{6.48}\pm0.07$ & $~~3.49\pm0.09$\\
 &  & \modelbname & $~~6.49\pm0.06$ & $~~3.53\pm0.08$\\ \hline
 \end{tabular}}
\end{table}

\paragraph{Experimental Setup}
Although our exposition in Section \ref{sec:sparse-moe} was for a single-task setting, the same gate can also be used in multi-task learning --- multi-task requires multi-gate MoE architecture \citep{ma2018modeling}, where each task has a separate trainable gate, but tasks have to select from a common set of experts. 
We briefly summarize the key aspects for each dataset. For MovieLens/Books/Jester we have two tasks: classification task predicts whether user watches/read/rates a particular movie/book/joke, regression problem predicts user's rating. Loss is the convex combination of the two binary cross-entropy (for classification) and mean squared error (for regression) with task weights: $\{\alpha, 1-\alpha\}$. We separately present results for two different $\alpha$'s: $\alpha \in \{0.1, 0.9\}$.
For MultiMNIST/MultiFashionMNIST, there are two multi-class classification tasks, which are equally weighted.
For CelebA, there are 10 binary classification problems, which are equally weighted.
Lastly, for Digits dataset, we have a multi-class single-task classification cross-entropy objective. 
Full details about datasets and MoE architectures are in Supplement Section \ref{supp-sec:appendix-recommender-vision}.

We used Adam for optimization, and we tuned the key hyperparameters using random grid search.
Note that for \modeldname, \modelbname~and \modelcname, we only allocate a very small portion of the epochs (1-10) for permutation learning.
Full details about the hyperparameter tuning are given in Supplement Section \ref{supp-sec:appendix-recommender-vision}.

\subsubsection{Performance of \modelaname~and \modelbname}
\label{sec:results-on-recommender-and-vision-datasets}
In Tables \ref{tab:comet-rec-sys} and \ref{tab:comet-image}, we report the (average) test loss and the average number of selected experts per sample across multiple recommender and vision datasets. The results indicate that \modelaname~ and \modelbname~lead on many datasets, outperforming popular state-of-the-art gating methods e.g., Hash-r, Top-k and DSelect-k in test loss.
Our proposed gate \modelaname~can outperform standard routing techniques (without local search). Even without local search, \modelaname~is getting relatively good solutions. We hypothesize that the good performance of \modelaname~is due to a combination of factors including differentiability, and $k$-decision trees formulation. With local search, \modelbname~can sometimes further enhance solution quality. 
We also provide task-specific metrics (AUC/Accuracy/MSE) in Tables \ref{tab:task-specific-metrics} in Appendix \ref{supp-sec:additional-results}.
We observe \modelbname~can improve AUC by up to $13\%$ over Hash routing and Top-k, and $9\%$ over DSelect-k.
We observe that Top-k gate does not uniformly outperform the Softmax across multiple datasets. However, \modelcname~significantly improves the performance of Top-k across multiple datasets.
In fact with the permutation module, \modelcname~outperforms Softmax in all cases, so sparsity in gating seems to be beneficial on all these datasets.

\paragraph{Inference Sparsity} We see that \modelaname~and \modelbname~can sometimes lead to a smaller number of experts selected than that for Top-k. This leads to smaller number of FLOPs at inference time (see Appendix \ref{supp-sec:flops}).
For some settings, DSelect-k appears to arrive at a sparser selection than \modelbname; however, in these cases, DSelect-k loses significantly in terms of performance. We observed expert collapsing in DSelect-k in such cases. 

\paragraph{Timing Discussion} For cost complexity of \modelaname, please see Appendix \ref{supp-sec:cost-analysis}.
Additionally, we discuss the computational aspects of the local search in Appendix \ref{supp-sec:effect-of-local-search-on-computation}.  

\subsubsection{Reducing Hyperparameter Search with \modelaname}
\label{sec:reducing-tuning}
\begin{figure}[!b]
\centering
        \setlength{\tabcolsep}{4.0pt}       \begin{tabular}{cc}
        Books & Jester \\
        \includegraphics[width=0.425\columnwidth]{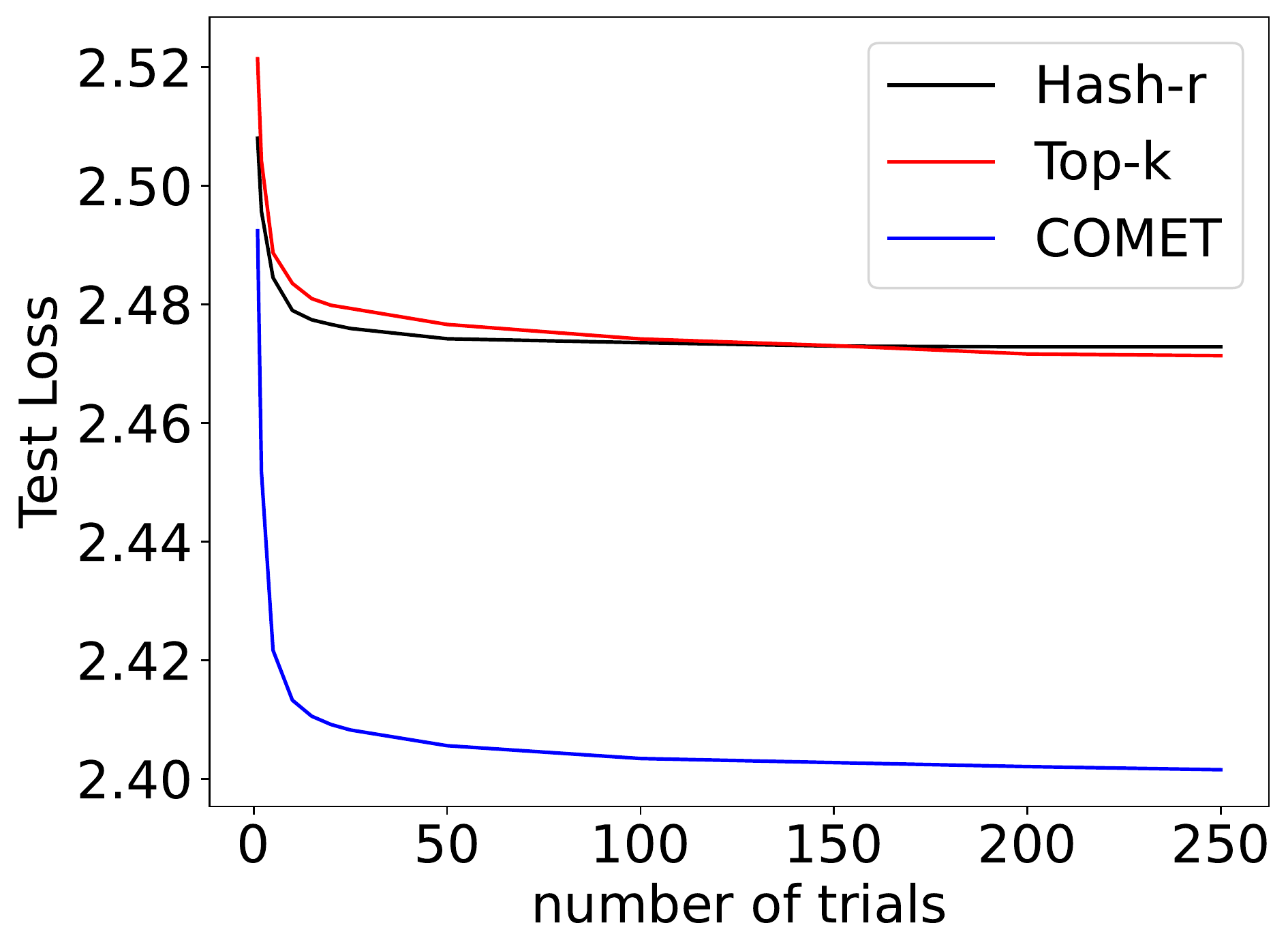} & 
        \includegraphics[width=0.425\columnwidth]{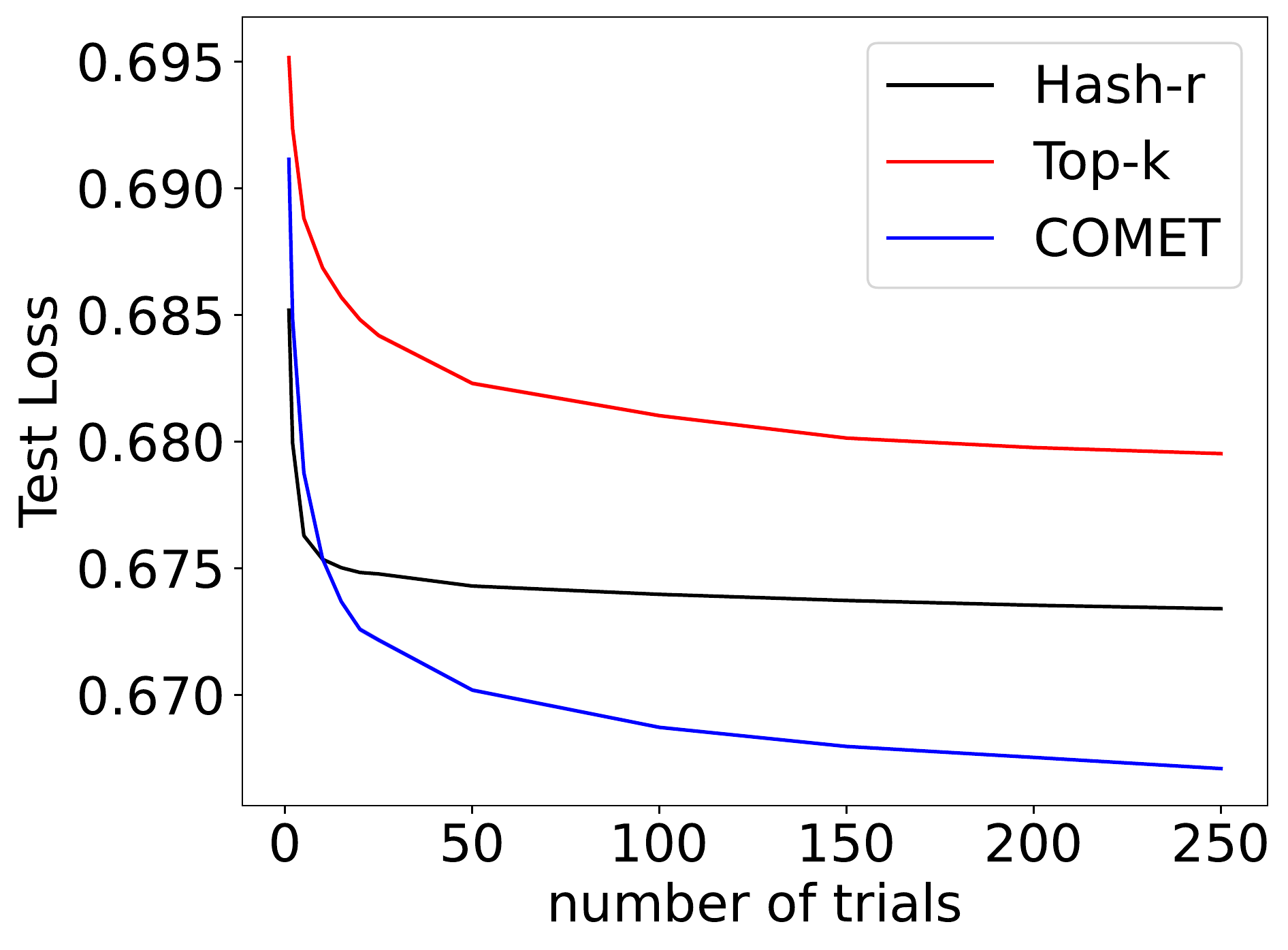} \\
        Digits & MovieLens \\        \includegraphics[width=0.425\columnwidth]{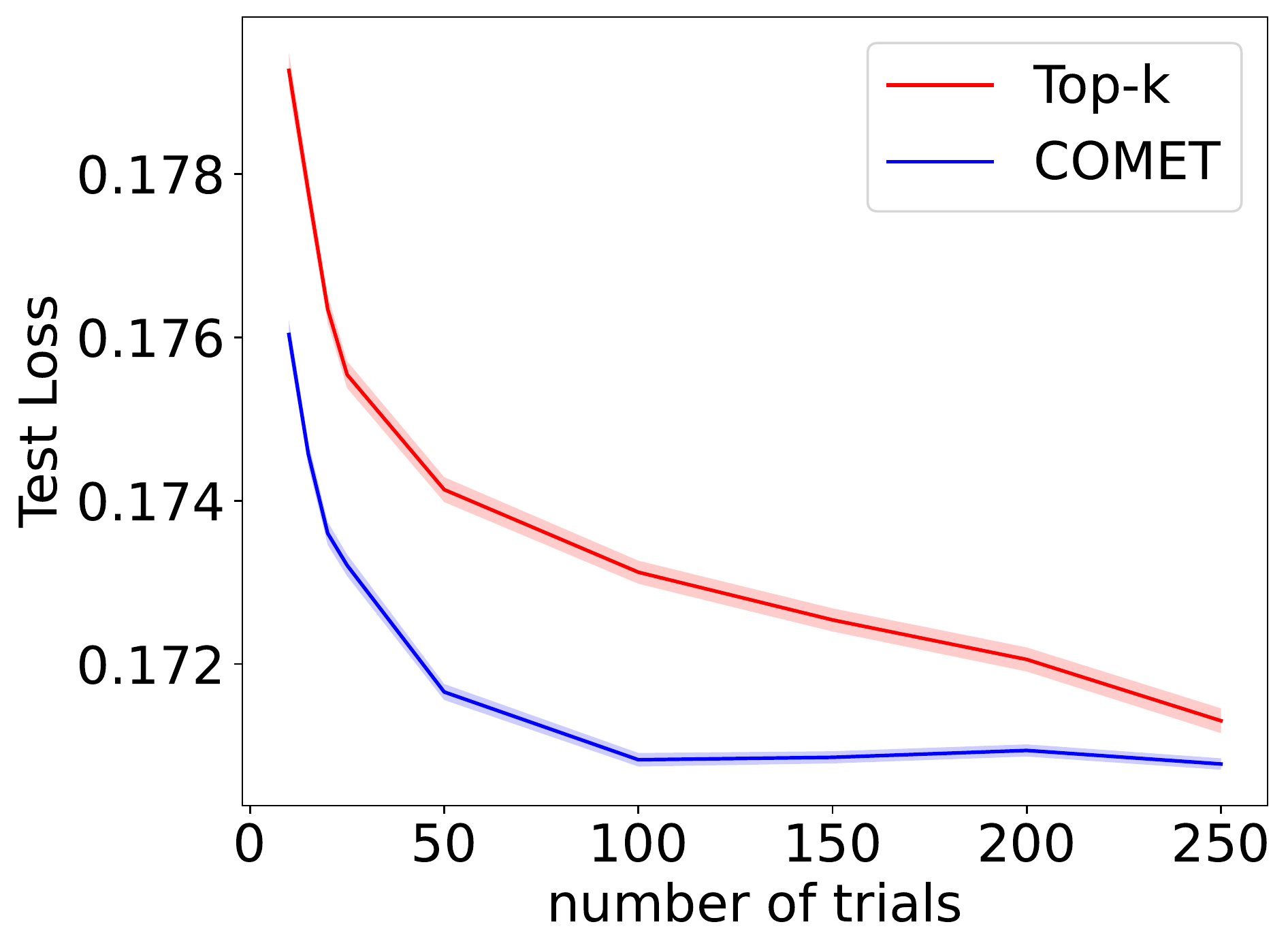} &
        \includegraphics[width=0.425\columnwidth]{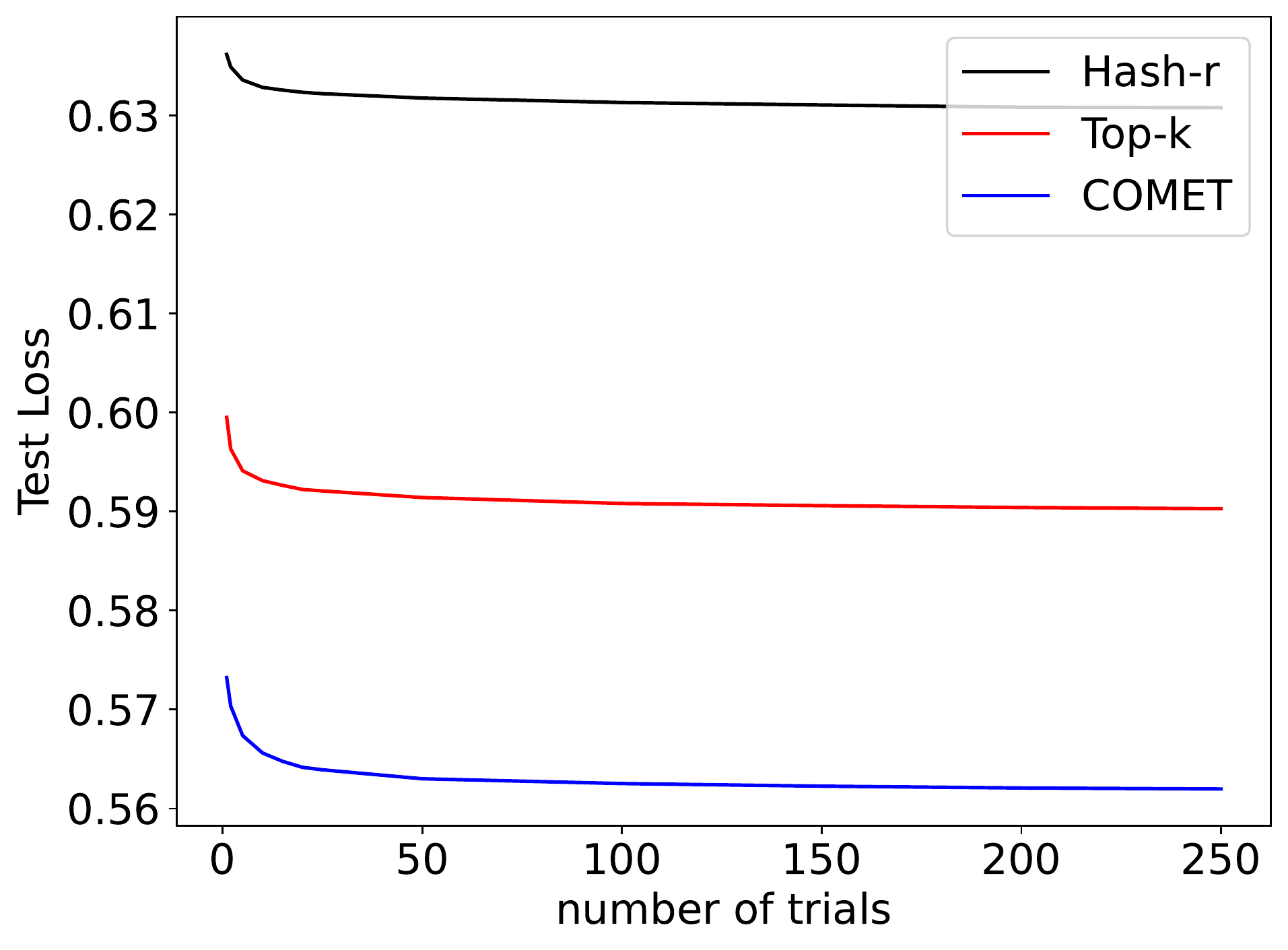} \\
        \end{tabular}
\caption{\emph{Sensitivity of \modelaname~to hyperparameter tuning}. \modelaname~can achieve the same level of performance as popular gates (e.g., Hash-r and Top-k) with significantly lesser number of hyperparameter trials. We see tuning reduction by $5\!\times\!-100\times$ for \modelaname~over Top-k and Hash routing.
} 
\label{fig:tuning-comet-vs-popular}
\end{figure}
Here, we study how our differentiable \modelaname ~gate (that performs dense-to-sparse training) can  be beneficial in terms of hyperparameter tuning over popular gates such as Hash routing and Top-k.
We perform a large set of tuning trials and perform a bootstrapping procedure (discussed in Appendix \ref{supp-sec:bootstrap-trials}) to see whether \modelaname~helps in reducing the hyperparameter tuning overload.
\modelaname~can achieve the same level of performance as popular gates with much lesser number of hyperparameter trials.
This indicates that \modelaname~is not too heavily dependent on a very restricted set of hyperparameter values.
We visualize this for various datasets in Fig. \ref{fig:tuning-comet-vs-popular}.
We see tuning reduction by a factor of $5\!\times\!-100\times$ for \modelaname~over popular gates.

\subsubsection{Effect of Local Search on Hyperparameter Tuning}
\label{sec:bootstrap-trials}
\begin{figure}[!b]
        \setlength{\tabcolsep}{4.0pt}       \begin{tabular}{cc}
        MovieLens ($\alpha=0.1$) & 
        MovieLens ($\alpha=0.9$) \\
        \includegraphics[width=0.40\columnwidth]{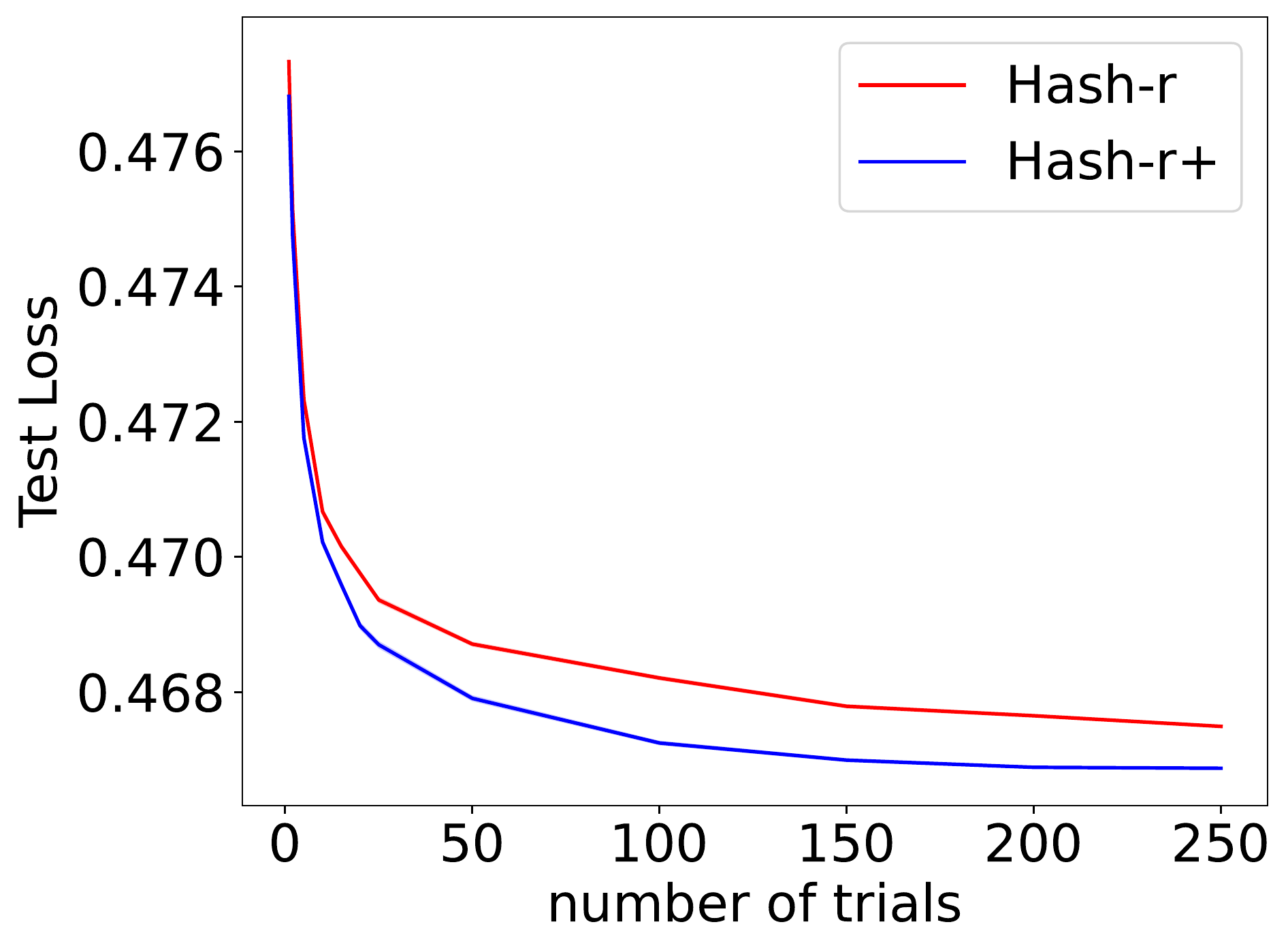} &   
        \includegraphics[width=0.40\columnwidth]{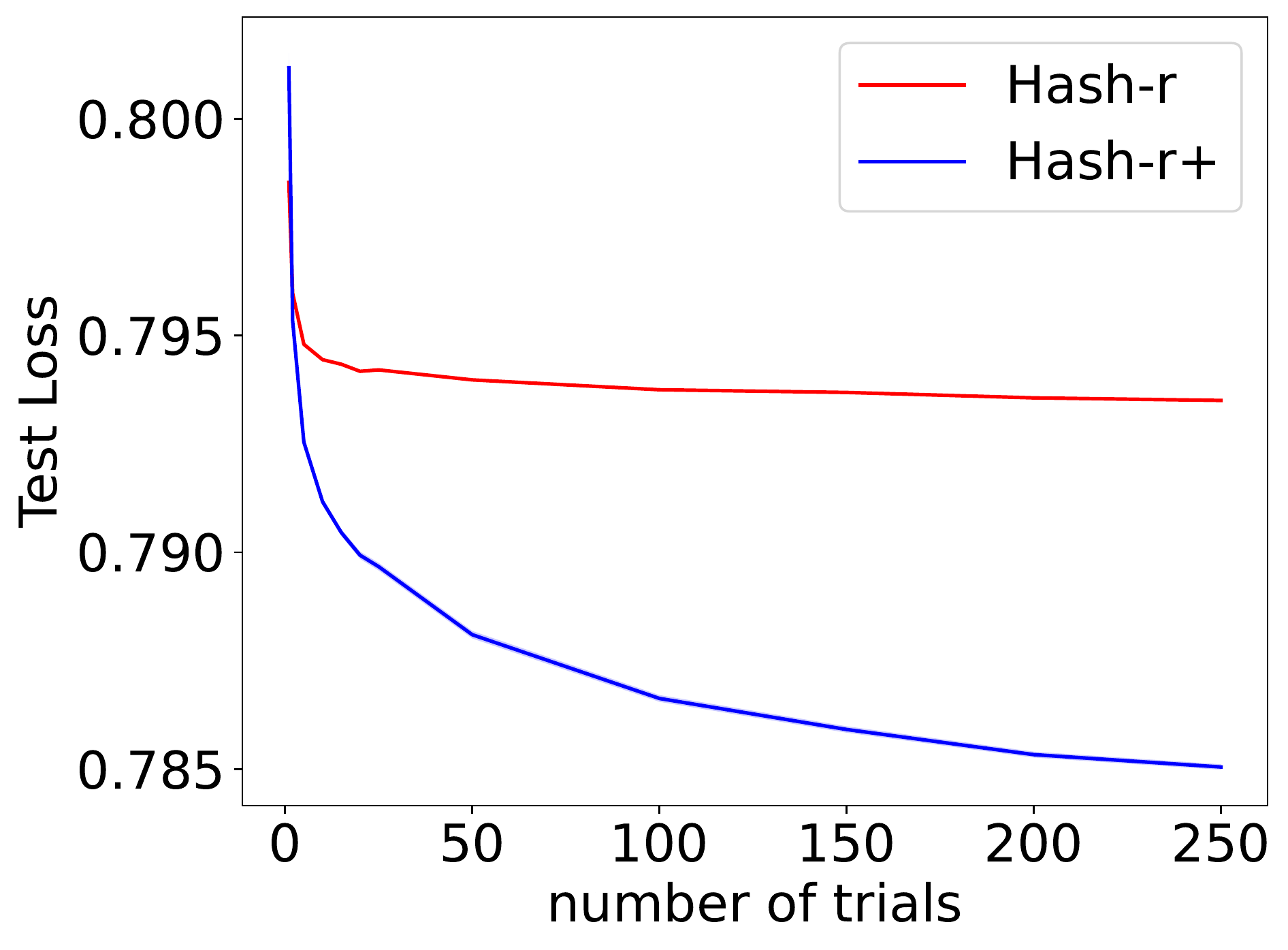} \\   
        
        \includegraphics[width=0.40\columnwidth]{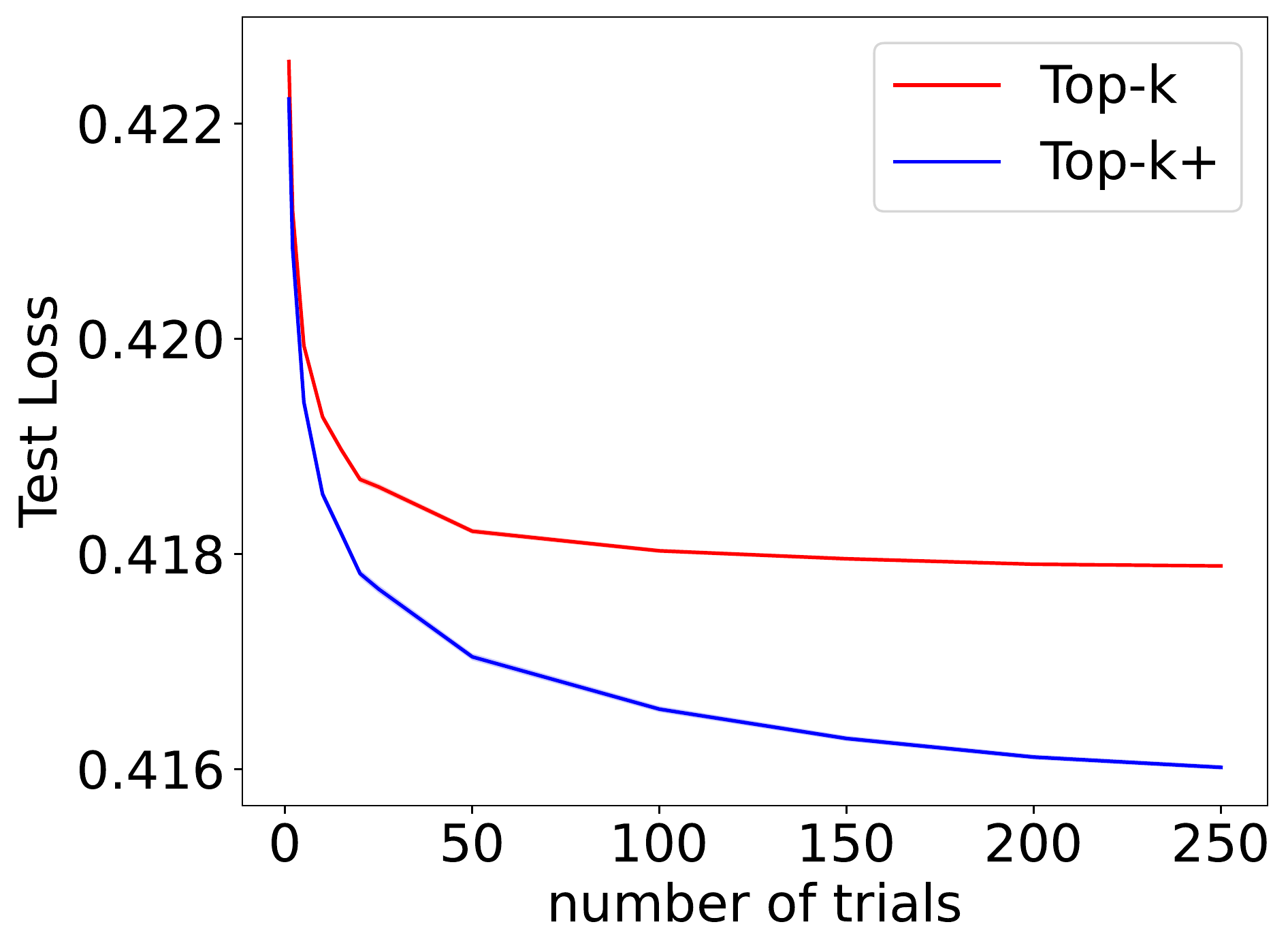} & 
        \includegraphics[width=0.40\columnwidth]{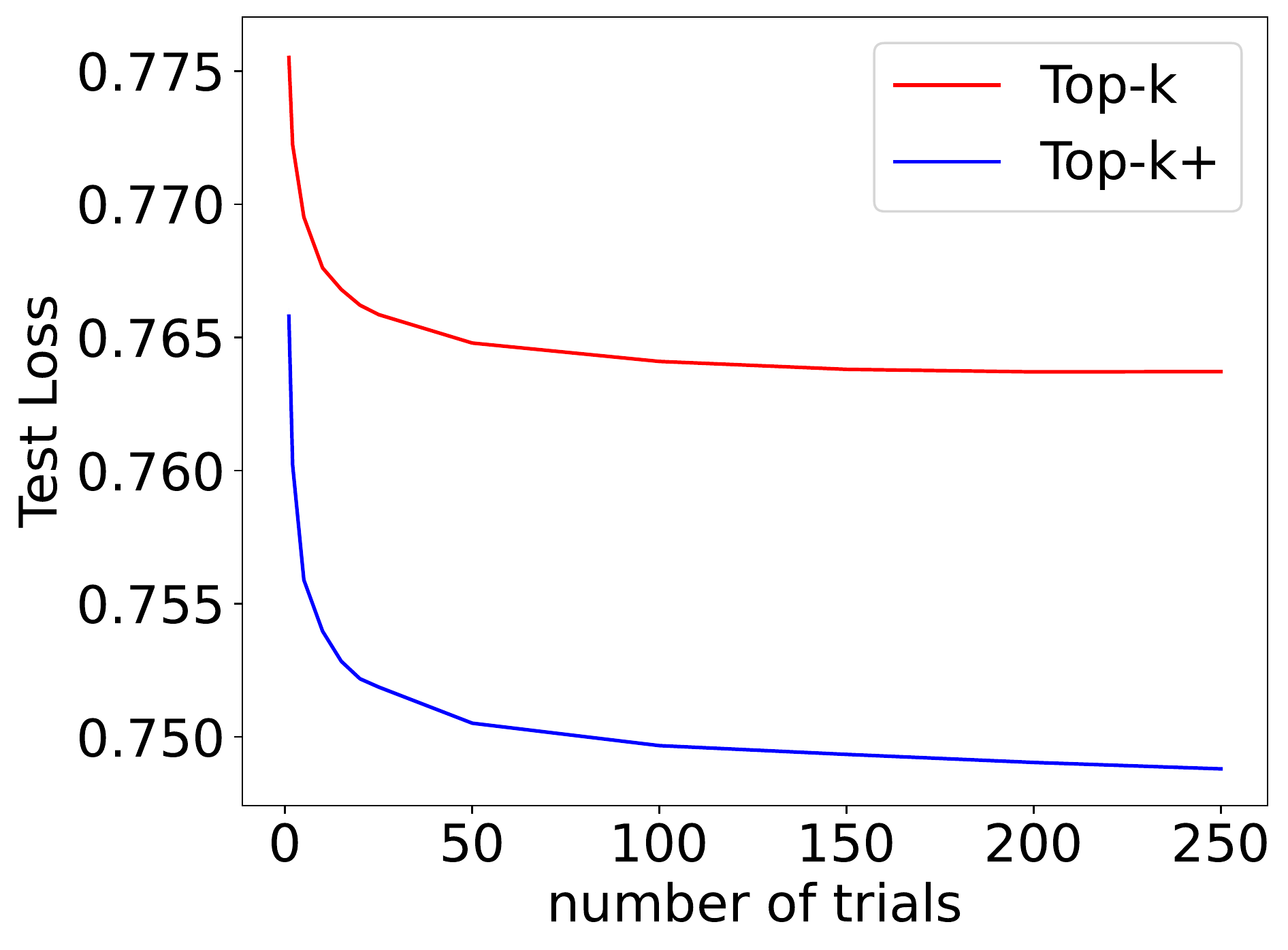} \\   

        \includegraphics[width=0.40\columnwidth]{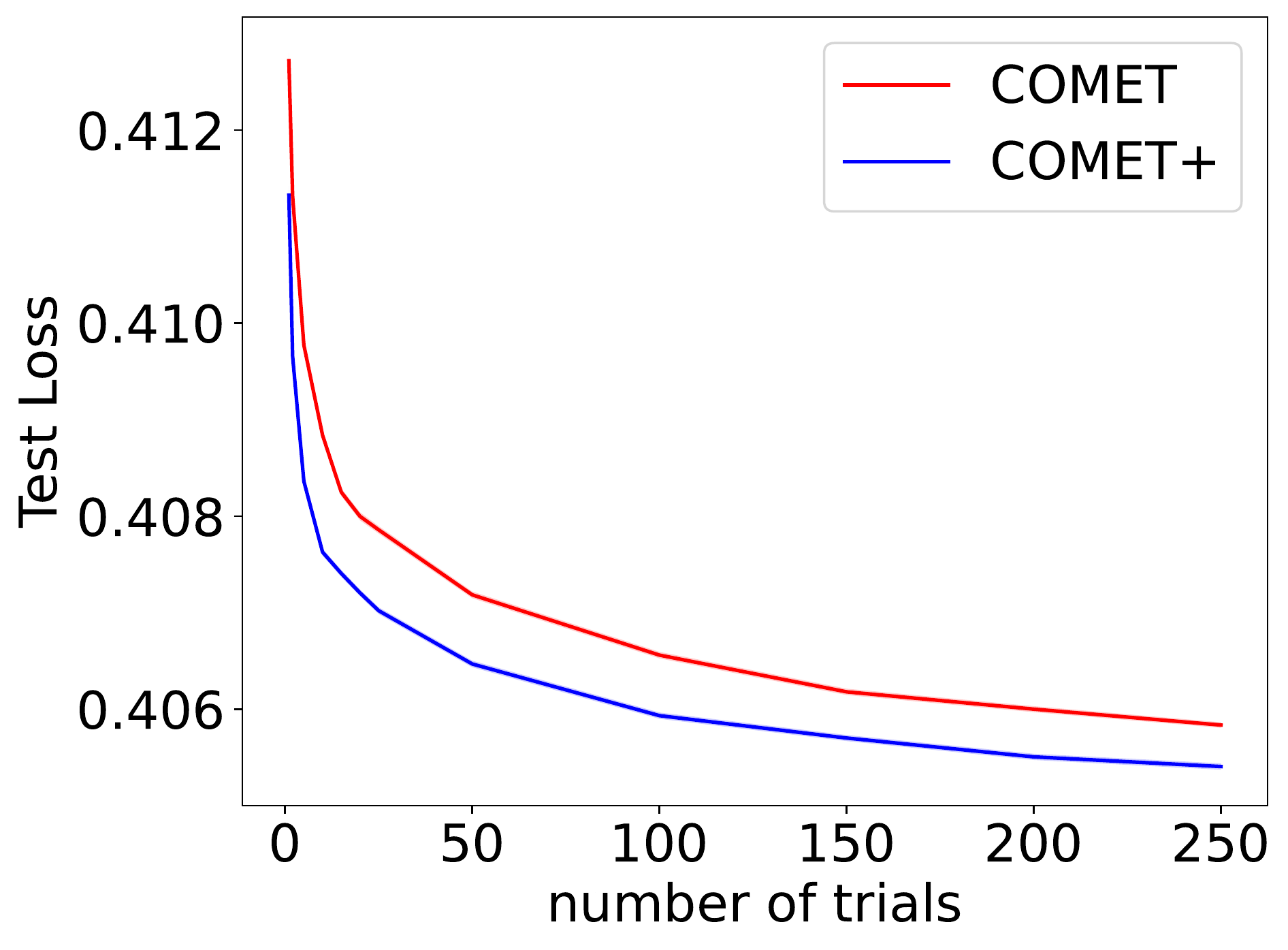} & 
        \includegraphics[width=0.40\columnwidth]{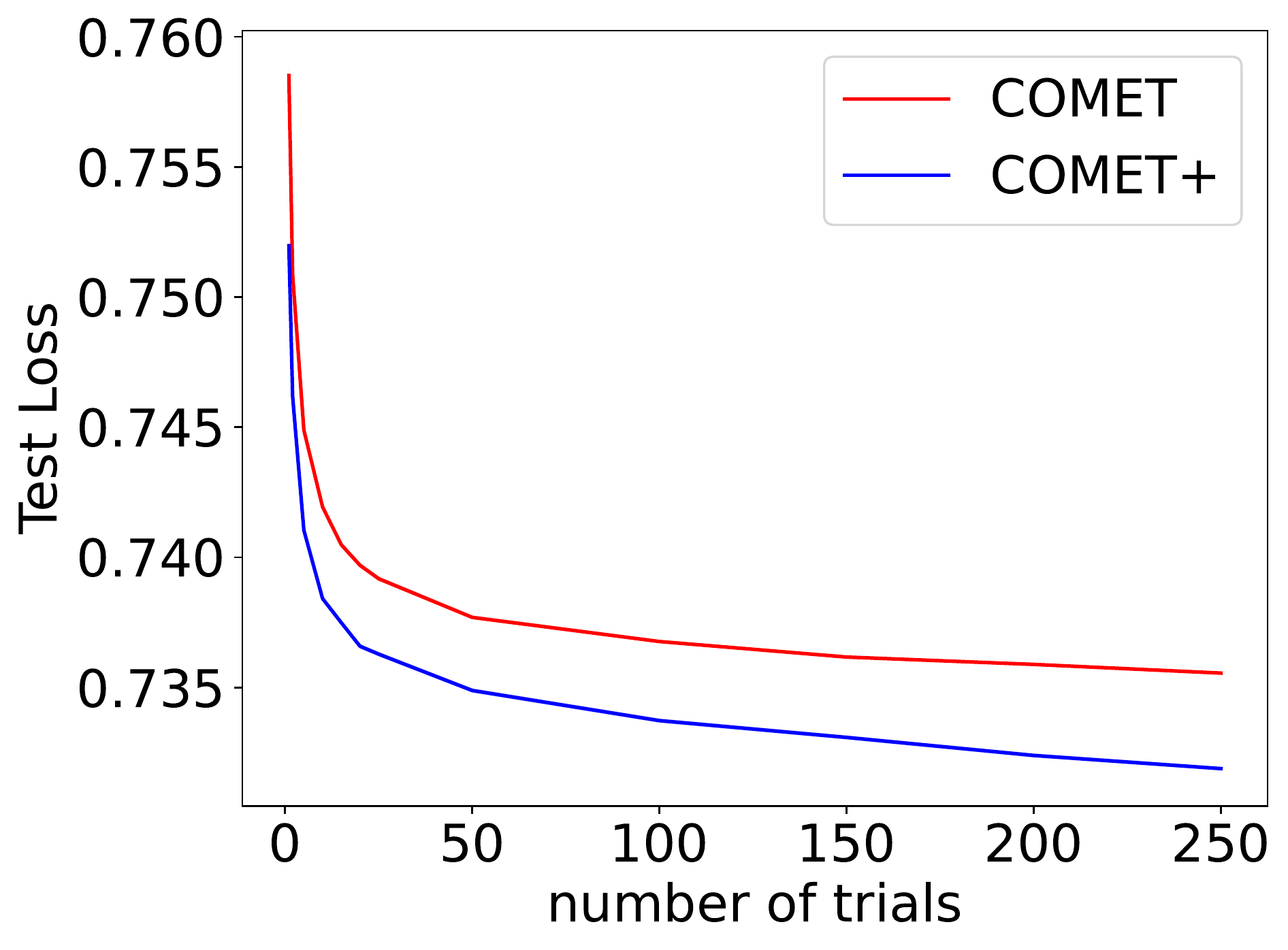}       
        \end{tabular}
\caption{\emph{Effect of local search on hyperparameter tuning}. Comparison of \modeldname~vs Hash-r, \modelcname~vs Top-k and \modelbname~vs \modelaname~on MovieLens with two different task weight settings. Local search appears to achieve the same level of performance with much lesser number of hyperparameter trials. We see tuning reduction by a factor of $3\!\times\!-100\times$ for \modeldname, $20\!\times\!-100\times$ for \modelcname~and $2\!\times\!-5\times$ for \modelbname.
} 
\label{fig:effect-of-permutation}
\end{figure}
Here, we study how local search can be beneficial in terms of hyperparameter tuning. We study this effect for Hash-r, Top-k and \modelaname.
We visualize this in Fig. \ref{fig:effect-of-permutation} for MovieLens for both \modeldname, \modelcname and \modelbname.
We observe that we can achieve comparable performance with much smaller number of trials. We see tuning reduction by a factor of $3\!\times\!-100\times$ for \modeldname, $20\!\times\!-100\times$ for \modelcname~and $2\!\times\!-5\times$ for \modelbname. 
This suggests that permutation-based local search helps escape out of bad initializations.
Such favorable properties of local search in terms of reducing the hyperparameter tuning load for existing gates can be beneficial for Large Language Models.

\subsection{Experiments on NLP Tasks}
\label{sec:moebert}
In this section, we consider a setting where a pretrained large model (non-MoE based) needs to be distilled for a more efficient inference while preserving or improving the best performance.
Following \cite{Zuo2022moebert}, we study a distillation setting, where BERT \citep{Devlin2018} is distilled into its Sparse-MoE based variant.
Specifically, the FFN layers are replaced with MoE layers --- this can result in a $\sim\!\!2\times$ smaller number of (effective) parameters with per-sample sparse routing (for $k=1$), thus allowing for more efficient inference.

Following \cite{Zuo2022moebert}, we use an importance-weight guided distillation strategy: (i) Finetune BERT on a downstream task.
(ii) Compute importance weights in FFN layers to construct an MoE-based variant of BERT.
(iii) Distill BERT into MoE-based variant on the downstream task with a layer-wise discrepancy loss. 
\cite{Zuo2022moebert} used Hash routing in their MoEBERT model.
We propose \modelename~(MoE based BERT model with \modelaname / \modelbname~gating) and evaluate the performance on the GLUE benchmarks \citep{Wang2019} and SQuAD benchmark \citep{rajpurkar-etal-2018-know}.
More details about the benchmarks are given in Supplement Section \ref{supp-sec:nlp-datasets}.

\paragraph{Implementation} We implemented \modelename~in HuggingFace \citep{Wolf2020} and adapted the codebase of \cite{Zuo2022moebert}. Unlike Hash routing, our gates can also cater to $k\geq1$. However, for consistent comparison in terms of inference, we set $k=1$.
Tuning details are outlined in Supplement Section \ref{supp-sec:moebert-tuning}. Code for \modelename~is available at \url{https://github.com/mazumder-lab/COMET-BERT}.

\paragraph{Results}
We report the performance metrics in Table \ref{tab:moebert} for 7 GLUE datasets and SQuAD dataset. 
\modelename~ outperforms MoEBERT in 5/7 benchmarks on GLUE datasets. \modelename~ also outperform MoEBERT significantly on SQuADv2.0. Notably, in 5 of these datasets (CoLA, MRPC, QNLI and MNLI, SQuAD v2.0), \modelename~ achieves SOTA performance when distilling BERT, (when compared with all distillation methods in literature with same number of effective parameters for inference). 

\begin{table}[!h]
\centering
\small
\caption{Performance metrics on the GLUE and SQuAD development sets. Models are trained without data augmentation. Both models have 66M (effective) parameters for inference.}
\label{tab:moebert}
\resizebox{\columnwidth}{!}{
\setlength{\tabcolsep}{2.0pt}
\begin{tabular}{l|ccccccc|c|}
\cline{2-9}
 & \multicolumn{7}{c|}{\textbf{GLUE}} & \textbf{SQuAD} \\ \cline{2-9} 
 & \multicolumn{1}{c|}{\begin{tabular}[c]{@{}c@{}}\textbf{RTE}\\ Acc\end{tabular}} & \multicolumn{1}{c|}{\begin{tabular}[c]{@{}c@{}}\textbf{CoLA}\\ Mcc\end{tabular}} & \multicolumn{1}{c|}{\begin{tabular}[c]{@{}c@{}}\textbf{MRPC}\\ F1\end{tabular}} & \multicolumn{1}{c|}{\begin{tabular}[c]{@{}c@{}}\textbf{SST-2}\\ Acc\end{tabular}} & \multicolumn{1}{c|}{\begin{tabular}[c]{@{}c@{}}\textbf{QNLI}\\ Acc\end{tabular}} & \multicolumn{1}{c|}{\begin{tabular}[c]{@{}c@{}}\textbf{QQP}\\ F1/Acc\end{tabular}} & \begin{tabular}[c]{@{}c@{}}\textbf{MNLI}\\ m/mm\end{tabular} & \begin{tabular}[c]{@{}c@{}}\textbf{v2.0}\\ F1/EM\end{tabular} \\ \hline
\multicolumn{1}{|l|}{\textbf{MOEBERT}\tablefootnote{Numbers are based on re-run of the official codebase (\url{https://github.com/SimiaoZuo/MoEBERT}) with hash routing with the optimal hyperparameters reported in \citep{Zhou2022}.}} & \multicolumn{1}{c|}{70.8} & \multicolumn{1}{c|}{55.4} & \multicolumn{1}{c|}{91.0} & \multicolumn{1}{c|}{\textbf{93.2}} & \multicolumn{1}{c|}{90.9} & \multicolumn{1}{c|}{\textbf{88.5}/91.4} & 84.7 & 76.8/73.6 \\ \hline
\multicolumn{1}{|l|}{\textbf{\modelename}} & \multicolumn{1}{c|}{\textbf{71.1}} & \multicolumn{1}{c|}{\textbf{57.0}} & \multicolumn{1}{c|}{\textbf{91.3}} & \multicolumn{1}{c|}{93.0} & \multicolumn{1}{c|}{\textbf{91.2}} & \multicolumn{1}{c|}{88.4/-----} & \textbf{85.5} & \textbf{78.4}/\textbf{75.3} \\ \hline
\end{tabular}
}
\end{table}

\section{Conclusion}
In summary, we propose two new approaches for improving routing in Sparse-MoE. 
First, we introduce a new differentiable gate \modelaname, which relies on a novel tree-based sparse expert selection mechanism.
\modelaname~ allows optimization with first-order methods, offers explicit control over the number of experts to select, allows (partially) conditional training and sparse inference.
Second, in this work, we argue that combinatorial nature of expert selection in Sparse-MoE makes sparse routing optimization challenging with first-order methods.
Thus, we propose a new local search method that can help any gate including ours (\modelaname) escape ``bad'' initializations. 
Our large-scale experiments on recommender systems, vision and natural language processing tasks show \modelaname~ and \modelbname: 
(i) achieve statistically significant improvements in prediction (up to $13\%$ improvement in AUC) and expert selection over popular sparse gates. 
(ii) reduce tuning up to a factor of $100\times$ to achieve the same level of performance as popular gates e.g., Top-k and Hash routing.
(iii) help \modelename~achieve state-of-the-art results for distilling BERT on GLUE and SQuAD benchmarks.

\begin{acks}
This research is supported in part by grants from Google and the Office of Naval Research (ONR N000142112841).
We thank Mathieu Sibue for his initial help in setting up the baseline experiments on MovieLens.
We thank Paul Theron for his help with the tuning experiments for \modelename~on NLP tasks.
The authors acknowledge the MIT SuperCloud and Lincoln Laboratory Supercomputing Center for providing HPC resources that have contributed to the research results reported within this paper. 
The authors also acknowledge Google Cloud Credits, which were used for distillation experiments on NLP tasks. 
\end{acks}

\bibliographystyle{ACM-Reference-Format}
\bibliography{references}

\newpage
\appendix
\section*{Appendix}
\section{Smooth-Step Activation Function}
\label{supp-sec:smooth-step}
In smooth routing, the internal nodes of a soft tree use an activation function $h$ in order to compute the routing probabilities.
The common choice for $h$ --- logistic (a.k.a. sigmoid) function in soft trees literature \citep{Jordan1993, Kontschieder2015,Frosst2017,Hehn2019} --- can not output exact zeros. 
This implies that any sample $x$ will reach every node in the tree with a positive probability. 
Thus, computing the output of mixture of experts will require computation over every expert.
The smooth-step activation function proposed in \cite{Hazimeh2020} can output exact zeros and ones, thus allowing for conditional computation.
Let $\gamma$ be a non-negative scalar parameter. 
The smooth-step function is: 
\begin{align}
    h(t) = \begin{cases}
    0 &\text{if $t \leq -\gamma/2$}\\
    -\frac{2}{\gamma^3}t^3 + \frac{3}{2\gamma}t + \frac{1}{2}&\text{if $\gamma/2 \leq t \leq \gamma/2$}\\
    1 &\text{if $t \geq \gamma/2$}
    \end{cases}
\end{align}
The smooth-step function is continuously differentiable, similar to the logistic function. Additionally, it performs hard routing, i.e., outside $[-\gamma/2, \gamma/2]$, the function produces exact zeros and ones.
For cardinality-obeying Sparse-MoE learning with trees (not studied in \citep{Hazimeh2020}), the goal for each tree is to perform hard routing for all samples. Therefore, we propose additional regularization on $\{h(w_q \cdot x),  1-h(w_q \cdot x)\}$ to encourage convergence of $v$ to a one-hot state (discussed in more detail in Section \ref{sec:cardinality-with-k-trees}).

\section{Proof for Proposition \ref{prop:sparse-simplex}}
\label{supp-sec:proof-for-proposition}
\begin{proof}
First off, it is straightforward to see that $g(x;\alpha,v)$ satisfies the simplex constraint in (\ref{eq:sparse-moe-sparse-simplex}):
\begin{equation}
    \sum_{i=1}^ng(x;\alpha,v)_i=\sum_{i=1}^n\frac{\sum_{j=1}^k\exp(\alpha_{i}^{(j)}(x)) \cdot v_i^{(j)}(x)}{\sum_{j=1}^k\sum_{i=1}^n\exp(\alpha_{i}^{(j)}(x))v_i^{(j)}(x)}=1.
\end{equation}

It remains to show that $\|g(x;\alpha,v)\|_0\leq k$ under the given conditions.
Recall the hierarchical binary encoding $v^{(j)}$ outputs a one-hot vector for each sample $x$ as the routing decision.
Let us denote by $\hat i_j$ the expert number selected by $j$-th tree. For now, let us assume that $\hat i_j$ are different for any $j$. Then, we have
\begin{equation}
    g(x;\alpha,v)_{\hat i_j}={\exp(\alpha_{\hat i_j}^{(j)}(x))}\Bigm/{\textstyle\sum\limits_{j\in[k]}\exp(\alpha_{\hat i_j}^{(j)}(x))},
\end{equation}
i.e., the weights are restricted on the $k$ experts selected by the $k$ trees, and the weights form a softmax activation of logits $\alpha_{\hat i_j}^{(j)}$'s.
Given that the $j$-th tree selects $\hat i_j$-th expert, we know that if $i\notin\{\hat i_1,\hat i_2,\ldots, \hat i_k\}$, $v_i^{(j)}=0$ for any $j$, and thus $g(x;\alpha,v)_i=0$, and this means that the support of $g(x;\alpha,v)$ is contained in the set $\{\hat i_1,\hat i_2,\ldots, \hat i_k\}$, which has at most $k$ elements. Therefore, the cardinality constraint in (\ref{eq:sparse-moe-sparse-simplex}) holds.
\end{proof}

\begin{figure}
\centering
    \includegraphics[width=0.9\columnwidth]{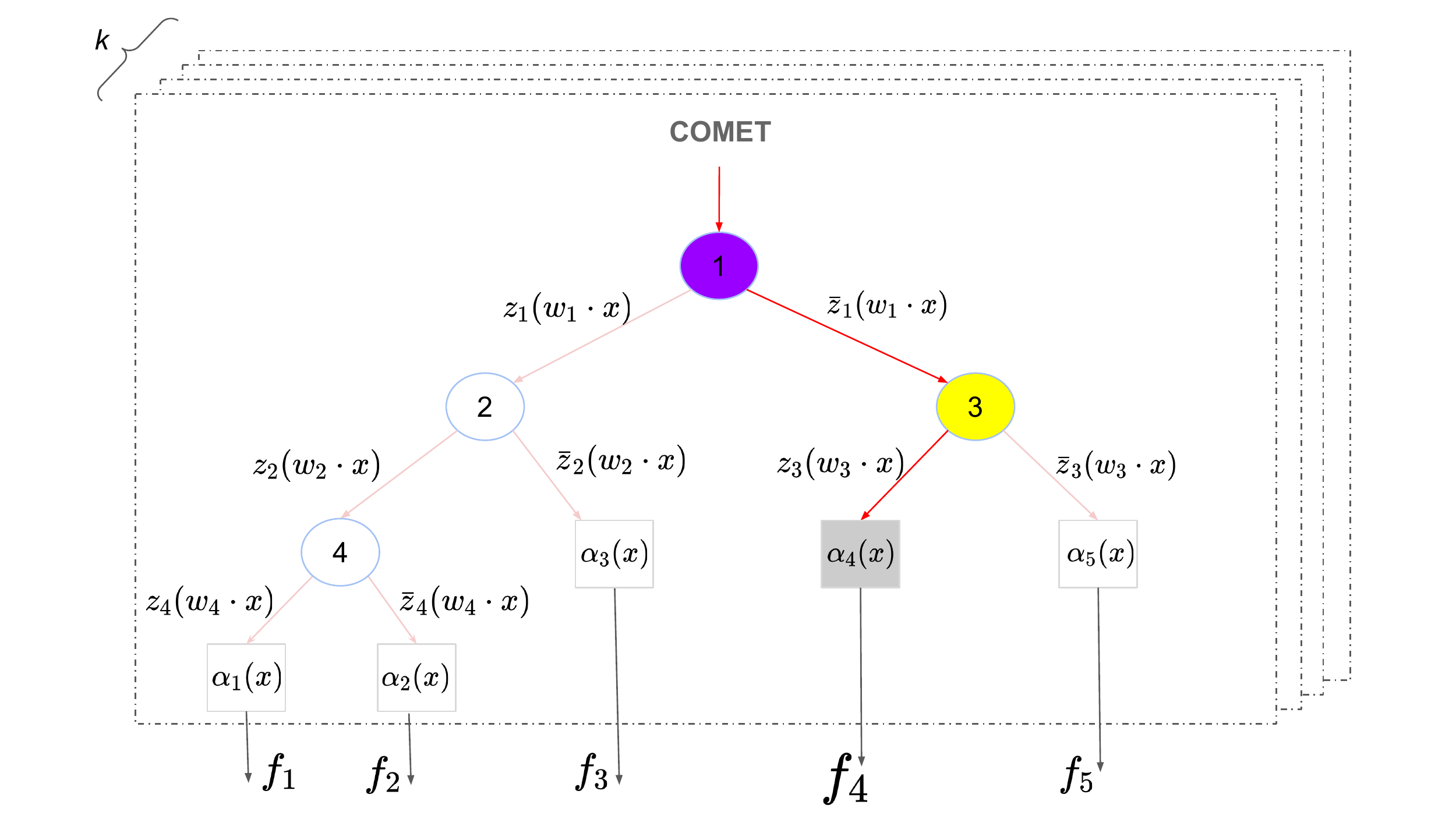}
\caption{\modelaname~for 5 (non-powers of 2) experts.} 
\label{fig:non-powers-of-2}
\end{figure}
\section{\modelaname~for non-powers of 2}
\label{supp-sec:non-powers-of-2}
\modelaname~has a natural way to cater to settings where number of experts are non-powers of 2. 
Recall we have $n$ experts.
Let $d=\lceil \log_2n\rceil,$ then $2^{d-1}<n\leq 2^d$.
Let $\mathcal{T}$ be a full binary tree with depth $d$ and $2^d$ leaf nodes.
We collapse $2(2^d-n)$ leaf nodes to their parents in the $(d-1)$-th level.
Each time we collapse two leaf nodes, we get a new node in the $(d-1)$-th level, and the total number of nodes reduce by one. 
Therefore, we get a tree with $n$ nodes, with $2^d-n$ nodes in the $(d-1)$-th level, and $2n-2^d$ nodes in the $d$-th level.
This is visualized in Figure \ref{fig:non-powers-of-2} for number of experts equal to 5.

\section{Timing Discussion}
\label{supp-sec:timing-discussion}
\subsection{Cost Complexity}
\label{supp-sec:cost-analysis}
We compare the cost of forward pass \modelaname~vs Top-k. We consider a single-task, $n$ experts ($i$th expert is $f_i$), $p$ features, desired sparsity $k \ll n$, and some shared layers.
\begin{table}[!h]
\centering
\caption{Time complexity of Top-k and \modelaname.}
\label{supp-tab:timing-complexity}
\resizebox{\columnwidth}{!}{
\setlength{\tabcolsep}{2.0pt}
    \begin{tabular}{c|ccc}
    \toprule
        Cost breakdown &  Top-k & \modelaname~(training) & \modelaname~(inference) \\
        \midrule
        Experts &  $kO(f_i)$ & $k_aO(f_i)$ & $k_a O(f_i)$\\
        Gate & $O(pn)$+$O(n+k\log n)$ & $O(kpn)$ & $O(kp\log n)$\\
        Shared layers &
        $O$(Shared Layers) & $O$(Shared Layers) & $O$(Shared Layers)\\
        \bottomrule
    \end{tabular}
}
\end{table}

\paragraph{Cost for Top-k} Here the desired sparsity ($k_a$) is always equal to the actual sparsity ($k$) during training and inference, i.e. $k_a\equiv k$. The cost of computing $g(\cdot)$ is given by $O(pn) + \mathcal{O}(k\log n+n)$. If the cost of computing expert $f_i$ is $O(f_i)$, then the cost of the full model is $k \mathcal{O}(f_i) + O(pn) + \mathcal{O}(k\log n+n) + 
\mathcal{O}(\text{shared~layers})$. Here the main cost is from how many $f_i$'s are evaluated per sample.\\

\paragraph{Cost for \modelaname} For \modelaname, the cost of the gate is $O(kpn)$.
The cost of the full model is $k_a \mathcal{O}(f_i) + O(kpn) + \mathcal{O}(\text{shared~layers})$. 
After a very few epochs (e.g., 4-5), when \modelaname~has reached the desired sparsity, we have $k_a \leq k$. 

Note that during inference, \modelaname~has smaller cost: 
All root-to-leaf paths in \modelaname~do not require evaluation, hence the gate cost reduces from $O(kpn)$ to $O(kp\log(n))$. Hence, due to fewer expert evaluations and smaller gate cost, \modelaname~is more efficient at inference time than Top-k.    
Also see Table \ref{supp-tab:flops} for FLOP counts at inference time for different gating methods.

\subsection{FLOPs Comparison}
\label{supp-sec:flops}
We compare the FLOP counts of different gating methods (across different models/datasets) to compare inference speed---see Table \ref{supp-tab:flops} below. Inference FLOPs per sample - number of floating point operations that a model performs per sample - is a standard measure to evaluate the inference speed for Sparse-MoE e.g., in \cite{Fedus2021} etc.

\begin{table}[!h]
\small
\centering
\captionsetup{width=.8\columnwidth}
\caption{\#FLOPs per-sample at inference time for \modelaname~against benchmarks (Softmax, Top-k, DSelect-k) across various datasets.}
\label{supp-tab:flops}
\resizebox{0.8\columnwidth}{!}{
\setlength{\tabcolsep}{20.0pt}
\begin{tabular}{|l|l|r|}
\hline
\textbf{Dataset} & \textbf{Model} & \textbf{FLOPs} \\ \hline
\multirow{4}{*}{Books} & Softmax & 1195K \\  
 & Top-k & 402K \\ 
 & DSelect-k & \textbf{214K} \\  
 & \modelaname & $326K$ \\ \hline
\multirow{4}{*}{MovieLens} & Softmax & 2255K \\  
 & Top-k & 413K \\  
 & DSelect-k & 399K \\  
 & \modelaname & \textbf{362K} \\ \hline
\multirow{4}{*}{MultiFashionMNIST} & Softmax & 7.49M \\  
 & Top-k & 3.03M \\ 
 & DSelect-k & \textbf{1.58M} \\  
 & \modelaname & 2.35M \\ \hline
\multirow{4}{*}{CelebA} & Softmax & 22.02M \\  
 & Top-k & 8.82M \\ 
 & DSelect-k & 5.64M \\  
 & \modelaname & \textbf{5.47M} \\ \hline
\end{tabular}}
\end{table}

We show that gates that learn sparse routing decisions per sample, e.g., Top-k, DSelect-k, \modelaname, significantly reduce the number of FLOPs ($3\!\times\!-6\times$) at inference time in comparison to dense gates e.g., Softmax. Additionally, we see that in all 4 cases, \modelaname~has smaller number of FLOPs ($1.1\!\times\!-1.6\times$) than the highly popular Topk gate. We also outperform DSelect-k in some cases in number of FLOPs. While in some cases, we have larger number of FLOPs than DSelect-k, our AUC is higher (up to 9\%) in these cases. 

\subsection{Effect of local search on computation}
\label{supp-sec:effect-of-local-search-on-computation}
\paragraph{Inference} Note that the permutation matrix is global and not sample specific.
At inference time, multiplying permutation matrix $\B{P}$ with $g(x)$ amounts to a reordering of the expert indices -- hence, additional cost for this permutation is negligible compared to evaluation of $f(x)$ and $g(x)$.

\paragraph{Training} In the first stage of \modelaname~training (a few epochs $\sim 5$), the training is dense (requiring all experts per sample). 
For \modelbname, we also learn the permutation matrix during this stage. There is a small additional computational cost: (a) permutation matrix of size $n \times n$, where $n$ is the number of experts, e.g., 16; (b) cost of Sinkhorn operator which constitutes row/column sum normalizations. This cost is marginal compared to the cost of evaluating the experts $f_i's$, each of which is an MLP/CNN. In the second stage of training, where the samples are being routed to a small $k(=2)$ subset of experts per-sample, there is no additional cost for \modelaname~vs \modelbname. To show an example, for MovieLens 200k, where we learn permutation matrix in first 5 epochs, the total time for 50 epochs (on 4 GPUs) is given by: 494s for \modelaname~and 496s for \modelbname. Note 50 epochs were sufficient to achieve convergence for both gates.

\section{Task-specific metrics corresponding to Tables \ref{tab:comet-rec-sys} and \ref{tab:comet-image}}
\label{supp-sec:additional-results}
We provide task-specific metrics for all recommender systems and image datasets in Table \ref{tab:task-specific-metrics}. We observe \modelbname~can give superior AUC performance by up to $13\%$ over Hash routing and Top-k, and $9\%$ over DSelect-k.  

\section{Bootstrapping Procedure for studying hyperparameter tuning}
\label{supp-sec:bootstrap-trials}
We performed 500 tuning trials and performed a bootstrapping procedure as outlined below:
\begin{itemize}[noitemsep,topsep=0pt,parsep=0pt,partopsep=0pt, leftmargin=*]
\item Randomly sample $s$ ($s \in \{1,2,5,10,15,\cdots,250\}$) trials from the bag of a larger set of 500 trials.
\item Find the trial with the best validation loss.
\item Compute the test loss for that trial. 
\item Repeat this exercise for 1000 times.
\item Compute the average test loss across the best selected trials.
\end{itemize}     

\begin{table}[!t]
\small
\centering
\caption{Test AUC/Accuracy/MSE for \modelbname~and benchmark gates on recommender systems and image datasets.}
\label{tab:task-specific-metrics}
\setlength{\tabcolsep}{12pt}
\resizebox{\columnwidth}{!}{\begin{tabular}{|llcc|}
\hline
\multicolumn{4}{|c|}{Recommender Systems} \\ \hline
\multicolumn{1}{|l|}{} & \multicolumn{1}{l|}{Gate} & \multicolumn{1}{c|}{\begin{tabular}[c]{@{}c@{}}Task 1 \\ (Test AUC)\end{tabular}} & \begin{tabular}[c]{@{}c@{}}Task 2\\ (Test MSE)\end{tabular} \\ \hline
\multicolumn{1}{|l|}{\multirow{5}{*}{\begin{tabular}[c]{@{}l@{}}Books\\ (alpha=0.1)\end{tabular}}} & \multicolumn{1}{l|}{Softmax} & \multicolumn{1}{c|}{56.70$\pm$0.16} & 2.6470$\pm$0.0016 \\
\multicolumn{1}{|l|}{} & \multicolumn{1}{l|}{Hash-r} & \multicolumn{1}{c|}{54.55$\pm$0.07} & 2.6791$\pm$0.0017 \\
\multicolumn{1}{|l|}{} & \multicolumn{1}{l|}{Top-k} & \multicolumn{1}{c|}{55.28$\pm$0.07} & 2.6783$\pm$0.0026 \\
\multicolumn{1}{|l|}{} & \multicolumn{1}{l|}{DSelect-k} & \multicolumn{1}{c|}{59.19$\pm$0.36} & 2.6667$\pm$0.0038 \\
\multicolumn{1}{|l|}{} & \multicolumn{1}{l|}{\modelbname} & \multicolumn{1}{c|}{\textbf{68.18}$\pm$0.24} & \textbf{2.6063}$\pm$0.0018 \\ \hline
\multicolumn{1}{|l|}{\multirow{5}{*}{\begin{tabular}[c]{@{}l@{}}Books\\ (alpha=0.9)\end{tabular}}} & \multicolumn{1}{l|}{Softmax} & \multicolumn{1}{c|}{77.85$\pm$0.01} & 2.6195$\pm$0.0019 \\
\multicolumn{1}{|l|}{} & \multicolumn{1}{l|}{Hash-r} & \multicolumn{1}{c|}{77.32$\pm$0.05} & 2.7152$\pm$0.0050 \\
\multicolumn{1}{|l|}{} & \multicolumn{1}{l|}{Top-k} & \multicolumn{1}{c|}{77.46$\pm$0.03} & 2.6942$\pm$0.0016 \\
\multicolumn{1}{|l|}{} & \multicolumn{1}{l|}{DSelect-k} & \multicolumn{1}{c|}{77.07$\pm$0.09} & 2.7581$\pm$0.0092 \\
\multicolumn{1}{|l|}{} & \multicolumn{1}{l|}{\modelbname} & \multicolumn{1}{c|}{\textbf{77.95}$\pm$0.02} & \textbf{2.6158}$\pm$0.0021 \\ \hline
\multicolumn{1}{|l|}{\multirow{5}{*}{\begin{tabular}[c]{@{}l@{}}MovieLens\\ (alpha=0.9)\end{tabular}}} & \multicolumn{1}{l|}{Softmax} & \multicolumn{1}{c|}{90.92$\pm$0.01} & 0.7585$\pm$0.0005 \\
\multicolumn{1}{|l|}{} & \multicolumn{1}{l|}{Hash-r} & \multicolumn{1}{c|}{88.95$\pm$0.02} & 0.8065$\pm$0.0004 \\
\multicolumn{1}{|l|}{} & \multicolumn{1}{l|}{Top-k} & \multicolumn{1}{c|}{91.25$\pm$0.01} & 0.7635$\pm$0.0008 \\
\multicolumn{1}{|l|}{} & \multicolumn{1}{l|}{DSelect-k} & \multicolumn{1}{c|}{91.65$\pm$0.02} & 0.7455$\pm$0.0006 \\
\multicolumn{1}{|l|}{} & \multicolumn{1}{l|}{\modelbname} & \multicolumn{1}{c|}{\textbf{91.70}$\pm$0.01} & \textbf{0.7437}$\pm$0.0006 \\ \hline
\multicolumn{1}{|l|}{\multirow{5}{*}{\begin{tabular}[c]{@{}l@{}}MovieLens\\ (alpha=0.1)\end{tabular}}} & \multicolumn{1}{l|}{Softmax} & \multicolumn{1}{c|}{85.50$\pm$0.03} & 0.7867$\pm$0.0029 \\
\multicolumn{1}{|l|}{} & \multicolumn{1}{l|}{Hash-r} & \multicolumn{1}{c|}{84.27$\pm$0.07} & 0.8279$\pm$0.0003 \\
\multicolumn{1}{|l|}{} & \multicolumn{1}{l|}{Top-k} & \multicolumn{1}{c|}{87.12$\pm$0.04} & 0.8005$\pm$0.0004 \\
\multicolumn{1}{|l|}{} & \multicolumn{1}{l|}{DSelect-k} & \multicolumn{1}{c|}{\textbf{88.16$\pm$0.07}} & 0.7734$\pm$0.0005 \\
\multicolumn{1}{|l|}{} & \multicolumn{1}{l|}{\modelbname} & \multicolumn{1}{c|}{88.02$\pm$0.01} & \textbf{0.7707}$\pm$0.0005 \\ \hline
\multicolumn{1}{|l|}{\multirow{5}{*}{\begin{tabular}[c]{@{}l@{}}Jester\\ (alpha=0.9)\end{tabular}}} & \multicolumn{1}{l|}{Softmax} & \multicolumn{1}{c|}{97.350$\pm$0.004} & 0.7460$\pm$0.0003 \\
\multicolumn{1}{|l|}{} & \multicolumn{1}{l|}{Hash-r} & \multicolumn{1}{c|}{97.323$\pm$0.003} & 0.7530$\pm$0.0003 \\
\multicolumn{1}{|l|}{} & \multicolumn{1}{l|}{Top-k} & \multicolumn{1}{c|}{97.346$\pm$0.004} & 0.7456$\pm$0.0006 \\
\multicolumn{1}{|l|}{} & \multicolumn{1}{l|}{DSelect-k} & \multicolumn{1}{c|}{97.361$\pm$0.004} & 0.7464$\pm$0.0005 \\
\multicolumn{1}{|l|}{} & \multicolumn{1}{l|}{\modelbname} & \multicolumn{1}{c|}{\textbf{97.362}$\pm$0.004} & \textbf{0.7439}$\pm$0.0006 \\ \hline
\multicolumn{1}{|l|}{\multirow{5}{*}{\begin{tabular}[c]{@{}l@{}}Jester\\ (alpha=0.1)\end{tabular}}} & \multicolumn{1}{l|}{Softmax} & \multicolumn{1}{c|}{97.22$\pm$0.01} & 0.7380$\pm$0.0003 \\
\multicolumn{1}{|l|}{} & \multicolumn{1}{l|}{Hash-r} & \multicolumn{1}{c|}{97.01$\pm$0.01} & 0.7301$\pm$0.0002 \\
\multicolumn{1}{|l|}{} & \multicolumn{1}{l|}{Top-k} & \multicolumn{1}{c|}{97.38$\pm$0.01} & 0.7412$\pm$0.0005 \\
\multicolumn{1}{|l|}{} & \multicolumn{1}{l|}{DSelect-k} & \multicolumn{1}{c|}{\textbf{97.45}$\pm$0.00} & 0.7273$\pm$0.0003 \\
\multicolumn{1}{|l|}{} & \multicolumn{1}{l|}{\modelbname} & \multicolumn{1}{c|}{\textbf{97.45}$\pm$0.00} & \textbf{0.7257}$\pm$0.0004 \\ \hline
\multicolumn{4}{|c|}{Image Datasets} \\ \hline
\multicolumn{1}{|l|}{} & \multicolumn{1}{l|}{Gate} & \multicolumn{2}{c|}{\begin{tabular}[c]{@{}c@{}}Test Accuracy\\ (Averaged across tasks)\end{tabular}} \\ \hline
\multicolumn{1}{|l|}{\multirow{4}{*}{MultiFashionMNIST}} & \multicolumn{1}{l|}{Softmax} & \multicolumn{2}{c|}{87.99$\pm$0.04} \\
\multicolumn{1}{|l|}{} & \multicolumn{1}{l|}{Top-k} & \multicolumn{2}{c|}{88.03$\pm$0.03} \\
\multicolumn{1}{|l|}{} & \multicolumn{1}{l|}{DSelect-k} & \multicolumn{2}{c|}{87.42$\pm$0.04} \\
\multicolumn{1}{|l|}{} & \multicolumn{1}{l|}{\modelbname} & \multicolumn{2}{c|}{\textbf{88.12}$\pm$0.04} \\ \hline
\multicolumn{1}{|l|}{\multirow{4}{*}{CelebA}} & \multicolumn{1}{l|}{Softmax} & \multicolumn{2}{c|}{83.84$\pm$0.15} \\
\multicolumn{1}{|l|}{} & \multicolumn{1}{l|}{Top-k} & \multicolumn{2}{c|}{83.95$\pm$0.17} \\
\multicolumn{1}{|l|}{} & \multicolumn{1}{l|}{DSelect-k} & \multicolumn{2}{c|}{83.53$\pm$0.06} \\
\multicolumn{1}{|l|}{} & \multicolumn{1}{l|}{\modelbname} & \multicolumn{2}{c|}{\textbf{84.27}$\pm$0.08} \\ \hline
\multicolumn{1}{|l|}{\multirow{4}{*}{Digits}} & \multicolumn{1}{l|}{Softmax} & \multicolumn{2}{c|}{93.53$\pm$0.12} \\
\multicolumn{1}{|l|}{} & \multicolumn{1}{l|}{Top-k} & \multicolumn{2}{c|}{95.34$\pm$0.04} \\
\multicolumn{1}{|l|}{} & \multicolumn{1}{l|}{DSelect-k} & \multicolumn{2}{c|}{95.41$\pm$0.03} \\
\multicolumn{1}{|l|}{} & \multicolumn{1}{l|}{\modelbname} & \multicolumn{2}{c|}{\textbf{95.45}$\pm$0.04} \\ \hline
\multicolumn{1}{|l|}{\multirow{4}{*}{MultiMNIST}} & \multicolumn{1}{l|}{Softmax} & \multicolumn{2}{c|}{98.01$\pm$0.03} \\
\multicolumn{1}{|l|}{} & \multicolumn{1}{l|}{Top-k} & \multicolumn{2}{c|}{98.01$\pm$0.02} \\
\multicolumn{1}{|l|}{} & \multicolumn{1}{l|}{DSelect-k} & \multicolumn{2}{c|}{98.03$\pm$0.02} \\
\multicolumn{1}{|l|}{} & \multicolumn{1}{l|}{\modelbname} & \multicolumn{2}{c|}{\textbf{98.07}$\pm$0.02} \\ \hline
\end{tabular}}
\end{table}

\newpage
\onecolumn
\section*{Supplementary Material for ``COMET: Learning Cardinality Constrained Mixture of Experts with Trees and Local Search''}
\setcounter{table}{0}
\renewcommand{\thetable}{S\arabic{table}}
\setcounter{figure}{0}
\renewcommand{\thefigure}{S\arabic{figure}}
\setcounter{equation}{0}
\renewcommand{\theequation}{S\arabic{equation}}
\setcounter{section}{0}
\renewcommand{\thesection}{S\arabic{section}}
\setcounter{footnote}{0}

\section{Additional Details for Section \ref{sec:experiments-recommender-systems}}
\label{supp-sec:appendix-recommender-vision}
\subsection{Datasets}

\paragraph{MovieLens}
MovieLens \cite{movielens2016} is a movie recommendation dataset containing records for $\sim4,000$ movies and $\sim6,000$ users. Following \cite{Wang2020}, for every user-movie pair, we construct two tasks. Task 1 is a binary classification problem for predicting whether the user will watch a particular movie. Task 2 is a regression problem to predict the user's rating (in $\{1, 2, \cdots, 5\}$) for a given movie. We use 1.6 million samples for training and 200, 000 for each of the validation and testing sets.

\paragraph{Jester}
Jester \cite{Goldberg2001} is a joke recommendation dataset containing records for $\sim74k$ users and $\sim100$ jokes. This gives a dataset of 7.4 million records.
Similar to MovieLens above, for every user-joke pair, we construct two tasks. 
Task 1 is a binary classification problem for predicting whether the user will rate a particular joke.
Task 2 is a regression problem to predict the user's rating (in $[-10,10]$) for a given joke.
We use 5.1 million samples for training and 1.1 million samples for each of the validation and testing sets.

\paragraph{Books}
Books \cite{Ziegler2005} is a book recommendation dataset containing records for $\sim105k$ users and $\sim340k$ books. We filter users and books with each atleast 5 records. This gives a subset of $18,960$ users and $31,070$ books. This gives a subset of 556,724 records.
Similar to MovieLens above, for every user-book pair, we construct two tasks. 
Task 1 is a binary classification problem for predicting whether the user will read a particular book.
Task 2 is a regression problem to predict the user's rating (in $\{1, 2, \cdots, 10\}$) for a given book.
We use 389,706 samples for training and 83,509 for each of the validation and testing sets.

\paragraph{Digits}
We use a mixture of MNIST \citep{Deng2012} and SVHN \citep{Netzer2011} datasets. MNIST is a database of $70,000$ handwritten digits. SVHN is a much harder dataset of $\sim600,000$ images obtained from house numbers in Google Street View images. We divided the dataset into training, validation and testing as follows:
MNIST (\#train: 50,000, \#validation: 10,000, \#test: 10,000) and SVHN (\#train: 480,420, \#validation: 75,000, \#test: 75,000). We combined the corresponding splits to get the train, validation and test sets for the mixture.

\paragraph{MultiMNIST/MultiFashionMNIST}
We consider multi-task variants of MNIST/MultiFashionMNIST \citep{Deng2012}. The datasets are constructed in a similar fashion as given in \cite{Sabour2017,Hazimeh2021}: (i) uniformly sample two images from MNIST and overlay them on top of each other, and (ii) shift one digit towards the top-left corner and the other digit towards the bottom-right corner (by 4 pixels in each direction).
This procedure leads to $36\times36$ images with some overlap between the digits. We consider two classification tasks: Task 1 is to classify the top-left item and Task 2 is to classify the bottom-right item. We use 100,000 samples for training, and $20,000$ samples for each of the validation and testing sets.

\paragraph{CelebA}
CelebA \citep{Liu2015} is a large-scale face attributes dataset with more than $200,000$ celebrity images, each with $40$ attribute annotations. The images in this dataset cover large pose variations and background clutter. We consider $10$ of the face attributes in a multi-task learning setting. We use $\sim160,000$ images for training, and $\sim 20,000$ for each of validation and testing.

\subsection{Architectures}
\label{supp-sec:architectures}
\paragraph{MovieLens}
We consider a multi-gate MoE architecture, where each task is associated with a separate gate.
The MoE architecture consists of a shared bottom subnetwork comprising two embedding layers (for users and movies). The 128-dimensional embeddings from both layers are concatenated and fed into an MoE Layer of 16 experts, where each expert is a ReLU-activated dense layer with 256 units, followed by a dropout layer (with a dropout rate of
0.5). For each of the two tasks, the corresponding convex combination of the experts is fed into a task-specific subnetwork. The subnetwork is composed of a dense layer (ReLU-activated with 256 units) followed by a single unit that generates the final output of the task.

\paragraph{Books/Jester}
We consider a multi-gate MoE architecture, where each task is associated with a separate gate.
The MoE architecture consists of a shared bottom subnetwork comprising two embedding layers (for users and books/jokes). The 64-dimensional embeddings from both layers are concatenated and fed into an MoE Layer of 9/16 experts, where each expert is a ReLU-activated dense layer with 128 units, followed by a dropout layer (with a dropout rate of
0.5). For each of the two tasks, the corresponding convex combination of the experts is fed into a task-specific subnetwork. The subnetwork is composed of a dense layer (ReLU-activated with 256 units) followed by a single unit that generates the final output of the task.

\paragraph{Digits}
We use a single-gate MoE with 8 experts. Each of the experts is a CNN that is composed (in order) of: (i) convolutional layer 1 (kernel size = 5, number of filters =
10, ReLU-activated) followed by max pooling, (ii) convolutional layer 2 (kernel size = 5, number of
filters = 20, ReLU-activated) followed by max pooling, and (iii) a ReLU-activated dense layer with 50 units.
The subnetwork specific to the prediction task is composed of a stack of 3 dense layers: the first two have 50 ReLU-activated units and the third has 10 units followed by a softmax.

\paragraph{MultiMNIST/MultiFashionMNIST}
We use a multi-gate MoE with 16/5 experts. Each of the  experts is a CNN that is composed (in order) of:
(i) convolutional layer 1 (kernel size = 5, \#filters = 10, ReLU-activated) followed by max pooling,
(ii) convolutional layer 2 (kernel size=5, \#filters = 20, ReLU-activated) followed by max pooling, and 
(iii) a sequence of 2 ReLU-activated dense layers with 50 units each.
The subnetwork specific to each of the 2 tasks is composed of a stack of 3 dense layers: the first two have 50 ReLU-activated units and the third has 10 units followed by a softmax.

\paragraph{CelebA}
We use a multi-gate MoE with 6 experts. Each of the  experts is a CNN that is composed (in order) of:
(i) convolutional layer 1 (kernel size = 3, \#filters = 4, ReLU-activated) followed by max pooling,
(ii) convolutional layer 2 (kernel size=3, \#filters = 4, ReLU-activated) followed by max pooling, (iii) convolutional layer 3 (kernel size=3, \#filters = 4, ReLU-activated) followed by max pooling, and 
(iv) convolutional layer 4 (kernel size=3, \#filters = 1, ReLU-activated) followed by max pooling, and 
(v) flatten layer. The subnetwork specific to each of the 2 tasks is composed of a dense layer followed by a sigmoid.

\subsection{Hyperparameters and Tuning}
We performed 500 tuning trials for each gate with a random search over the hyperparameter space described below (for each dataset).
For each gate, we selected Top $5\%$ of the trials based on validation loss.
We report the (average) test loss for the Top 5\% trials along with the standard errors in Tables \ref{tab:comet-rec-sys} and \ref{tab:comet-image}.

\paragraph{MovieLens}
\begin{itemize}[noitemsep,topsep=0pt,parsep=0pt,partopsep=0pt]
\item Learning Rates: Uniform in the log range $[5\times10^{-5}, 5\times10^{-4}]$ for Adam.
\item Batch-size: 512.
\item Epochs: 100 with early stopping (patience=25) based on validation set.
\item $\gamma$: Discrete uniform in the set $\{0.01, 0.1, 1, 5, 10\}$ for DSelect-k and \modelaname. $\gamma$ is fixed to $10$ for \modelbname.
\item Entropy: Discrete uniform in the set $\{0.05, 0.1, 0.5, 1, 5, 10\}$ for DSelect-k and \modelaname~and \modelbname.
\item Number of epochs for permutation learning: Discrete uniform in the set $\{1,\cdots,10\}$ for \modelbname~and \modelcname.
\item $\zeta$ (for permutation): $10^{-4}$
\item $n$ (number of experts): $16$.
\item $k$: 2 for all sparse (trainable) gates.
\item For Hash-r (and \modeldname), users are randomly pre-allocated to experts (similar to how words in vocabulary are pre-allocated randomly in LLMs)  
\item Number of tuning trials per gate: 500
\end{itemize}

\paragraph{Books}
\begin{itemize}[noitemsep,topsep=0pt,parsep=0pt,partopsep=0pt]
\item Learning Rates: Uniform in the log range $[5\times10^{-5}, 5\times10^{-4}]$ for Adam.
\item Batch-size: 2048.
\item Epochs: 100 with early stopping (patience=25) based on validation set.
\item $\gamma$: Discrete uniform in the set $\{0.1, 0.5, 1, 5, 10\}$ for DSelect-k and \modelaname. $\gamma$ is fixed to $0.5$ for \modelbname.
\item Entropy: Discrete uniform in the set $\{1, 5, 10, 50, 100\}$ for DSelect-k and \modelaname~and \modelbname.
\item Number of epochs for permutation learning: Discrete uniform in the set $\{1,\cdots,10\}$ for \modelbname~and \modelcname.
\item $\zeta$ (for permutation): $10^{-4}$
\item $n$ (number of experts): $9$.
\item $k$: 4 for all sparse (trainable) gates.
\item For Hash-r (and \modeldname), users are randomly pre-allocated to experts (similar to how words in vocabulary are pre-allocated randomly in LLMs)  
\item Number of tuning trials per gate: 500
\end{itemize}

\paragraph{Jester}
\begin{itemize}[noitemsep,topsep=0pt,parsep=0pt,partopsep=0pt]
\item Learning Rates: Uniform in the log range $[5\times10^{-5}, 5\times10^{-4}]$ for Adam.
\item Batch-size: 2048.
\item Epochs: 100 with early stopping (patience=25) based on validation set.
\item $\gamma$: Discrete uniform in the set $\{0.001, 0.02, 0.1, 1, 5, 10\}$ for DSelect-k and \modelaname. $\gamma$ is fixed to $0.01$ for \modelbname.
\item Entropy: Discrete uniform in the set $\{0.05, 0.1, 0.5, 1, 5, 10\}$ for DSelect-k and \modelaname~and \modelbname.
\item Number of epochs for permutation learning: Discrete uniform in the set $\{1,\cdots,10\}$ for \modelbname~and \modelcname.
\item $\zeta$ (for permutation): $10^{-4}$
\item $n$ (number of experts): $16$.
\item $k$: 2 for all sparse (trainable) gates.
\item For Hash-r (and \modeldname), users are randomly pre-allocated to experts (similar to how words in vocabulary are pre-allocated randomly in LLMs)  
\item Number of tuning trials per gate: 500
\end{itemize}

\paragraph{Digits}
\begin{itemize}[noitemsep,topsep=0pt,parsep=0pt,partopsep=0pt]
\item Learning Rates: Uniform in the log range $[1\times10^{-5}, 5\times10^{-4}]$ for Adam.
\item Batch-size: 512.
\item Epochs: 200 with early stopping (patience=25) based on validation set.
\item $\gamma$: Discrete uniform in the set $\{0.001, 0.01, 0.1, 1\}$ for DSelect-k and \modelaname. $\gamma$ is fixed to $0.001$ for \modelbname.
\item Entropy: Discrete uniform in the set $\{0.001, 0.005, 0.01, 0.05, 0.1, 0.5\}$ for DSelect-k and \modelaname~and \modelbname.
\item Number of epochs for permutation learning: Discrete uniform in the set $\{1,\cdots,10\}$ for \modelbname~and \modelcname.
\item $\zeta$ (for permutation): $10^{-4}$
\item $n$ (number of experts): $8$.
\item $k$: 2 for all sparse (trainable) gates.
\item Number of tuning trials per gate: 500
\end{itemize}

\paragraph{MultiMNIST}
\begin{itemize}[noitemsep,topsep=0pt,parsep=0pt,partopsep=0pt]
\item Learning Rates: Uniform in the log range $[1\times10^{-4}, 1\times10^{-3}]$ for Adam.
\item Batch-size: 512.
\item Epochs: 200 with early stopping (patience=25) based on validation set.
\item $\gamma$: Discrete uniform in the set $\{0.001, 0.01, 0.1, 1, 5, 10\}$ for DSelect-k and \modelaname. $\gamma$ is fixed to $0.01$ for \modelbname.
\item Entropy: Discrete uniform in the set $\{0.0001, 0.001,  0.01, 0.1, 1, 5, 10\}$ for DSelect-k and \modelaname~and \modelbname.
\item Number of epochs for permutation learning: Discrete uniform in the set $\{1,\cdots,10\}$ for \modelbname~and \modelcname.
\item $\zeta$ (for permutation): $10^{-4}$
\item $n$ (number of experts): $16$.
\item $k$: 4 for all sparse (trainable) gates.
\item Number of tuning trials per gate: 500
\end{itemize}

\paragraph{MultiFashionMNIST}
\begin{itemize}[noitemsep,topsep=0pt,parsep=0pt,partopsep=0pt]
\item Learning Rates: Uniform in the log range $[1\times10^{-4}, 1\times10^{-3}]$ for Adam.
\item Batch-size: 512.
\item Epochs: 200 with early stopping (patience=25) based on validation set.
\item $\gamma$: Discrete uniform in the set $\{0.001, 0.01, 0.1, 1, 5,\}$ for DSelect-k and \modelaname. $\gamma$ is fixed to $0.01$ for \modelbname.
\item Entropy: Discrete uniform in the set $\{0.001, 0.01, 0.1, 1, 5\}$ for DSelect-k and \modelaname~and \modelbname.
\item Number of epochs for permutation learning: Discrete uniform in the set $\{1,\cdots,10\}$ for \modelbname~and \modelcname.
\item $\zeta$ (for permutation): $10^{-4}$
\item $n$ (number of experts): $6$.
\item $k$: 2 for all sparse (trainable) gates.
\item Number of tuning trials per gate: 500
\end{itemize}

\paragraph{CelebA}
\begin{itemize}[noitemsep,topsep=0pt,parsep=0pt,partopsep=0pt]
\item Learning Rates: Uniform in the log range $[1\times10^{-4}, 1\times10^{-3}]$ for Adam.
\item Batch-size: 512.
\item Epochs: 100 with early stopping (patience=25) based on validation set.
\item $\gamma$: Discrete uniform in the set $\{0.001, 0.01, 0.1, 1, 5\}$ for DSelect-k and \modelaname. $\gamma$ is fixed to $5$ for \modelbname.
\item Entropy: Discrete uniform in the set $\{0.001, 0.01, 0.1, 1, 5\}$ for DSelect-k and \modelaname~and \modelbname.
\item Number of epochs for permutation learning: Discrete uniform in the set $\{1,\cdots,10\}$ for \modelbname~and \modelcname.
\item Entropy for permutation: $10^{-4}$
\item $k$: 2 for all sparse gates.
\item Number of tuning trials per gate: 100
\end{itemize}

\section{Additional Details for Section \ref{sec:moebert}}
\label{supp-sec:moebert}
\subsection{Datasets}
\label{supp-sec:nlp-datasets}
\paragraph{GLUE} General Language Understanding Evaluation (GLUE) benchmark \citep{Wang2019}, is a collection of natural language understanding tasks. Following previous works on model distillation,
we consider SST-2 \citep{Socher2013}, CoLA \citep{Warstadt2019}, MRPC \citep{Dolan2005}, STSB \citep{Cer2017}, QQP, and MNLI \citep{Williams2018} and exclude STS-B \citep{Cer2017} and WNLI \citep{Levesque2012} in the experiments. The datasets are briefly summarized below:
\begin{itemize}[noitemsep,topsep=0pt,parsep=0pt,partopsep=0pt]
\item SST-2 \citep{Socher2013} is a binary single-sentence classification task that classifies movie reviews to positive or negative;
\item CoLA \citep{Warstadt2019} is a linguistic acceptability task;
\item MRPC \citep{Dolan2005} is a paraphrase detection task;
\item QQP is a duplication detection task;
\item MNLI \citep{Williams2018}, QNLI \citep{Rajpurkar2016}, and RTE \citep{Dagan2006} are natural language inference tasks.
\end{itemize}
Dataset details are summarized in Table \ref{supp-tab:glue-datasets}.

\begin{table}[!h]
\centering
\caption{Summary of GLUE benchmark.}
\label{supp-tab:glue-datasets}
\setlength{\tabcolsep}{6pt}
\begin{tabular}{|lllllll|}
\hline
\multicolumn{1}{|l|}{Corpus} & \multicolumn{1}{l|}{Task} & \multicolumn{1}{l|}{\#Train} & \multicolumn{1}{l|}{\#Dev} & \multicolumn{1}{l|}{\#Test} & \multicolumn{1}{l|}{\#Label} & Metrics \\ \hline
\multicolumn{7}{|c|}{Single-Sentence Classification (GLUE)} \\ \hline
\multicolumn{1}{|l|}{CoLA} & \multicolumn{1}{l|}{Acceptability} & \multicolumn{1}{l|}{8.5k} & \multicolumn{1}{l|}{1k} & \multicolumn{1}{l|}{1k} & \multicolumn{1}{l|}{2} & Matthews correlation \\
\multicolumn{1}{|l|}{SST-2} & \multicolumn{1}{l|}{Sentiment} & \multicolumn{1}{l|}{67k} & \multicolumn{1}{l|}{872} & \multicolumn{1}{l|}{1.8k} & \multicolumn{1}{l|}{2} & Accuracy \\ \hline
\multicolumn{7}{|c|}{Pairwise Text Classification (GLUE)} \\ \hline
\multicolumn{1}{|l|}{MNLI} & \multicolumn{1}{l|}{Natural Language Inference} & \multicolumn{1}{l|}{393k} & \multicolumn{1}{l|}{20k} & \multicolumn{1}{l|}{20k} & \multicolumn{1}{l|}{3} & Accuracy \\
\multicolumn{1}{|l|}{RTE} & \multicolumn{1}{l|}{Natural Language Inference} & \multicolumn{1}{l|}{2.5k} & \multicolumn{1}{l|}{276} & \multicolumn{1}{l|}{3k} & \multicolumn{1}{l|}{2} & Accuracy \\
\multicolumn{1}{|l|}{QQP} & \multicolumn{1}{l|}{Paraphrase} & \multicolumn{1}{l|}{364k} & \multicolumn{1}{l|}{40k} & \multicolumn{1}{l|}{391k} & \multicolumn{1}{l|}{2} & Accuracy/F1 \\
\multicolumn{1}{|l|}{MRPC} & \multicolumn{1}{l|}{Paraphrase} & \multicolumn{1}{l|}{3.7k} & \multicolumn{1}{l|}{408} & \multicolumn{1}{l|}{1.7k} & \multicolumn{1}{l|}{2} & Accuracy/F1 \\
\multicolumn{1}{|l|}{QNLI} & \multicolumn{1}{l|}{Question Answering/Natural Language Inference} & \multicolumn{1}{l|}{108k} & \multicolumn{1}{l|}{5.7k} & \multicolumn{1}{l|}{5.7k} & \multicolumn{1}{l|}{2} & Accuracy \\ \hline
\end{tabular}
\end{table}

\paragraph{SQuAD}
We evaluate our sparse routing approaches on question answering dataset: SQuAD v2.0 \cite{rajpurkar-etal-2018-know}. This task is treated as a sequence labeling problem, where we predict the probability of each token being the start and end of the answer span. 
Statistics of the question answering dataset (SQuAD v2.0) are summarized in Table \ref{supp-tab:squad-datasets}.

\begin{table}[!h]
\centering
\caption{Summary of SQuAD benchmark.}
\label{supp-tab:squad-datasets}
\begin{tabular}{|lllll|}
\hline
\multicolumn{1}{|l|}{Corpus} & \multicolumn{1}{l|}{Task} & \multicolumn{1}{l|}{\#Train} & \multicolumn{1}{l|}{\#Dev} &  Metrics \\ \hline
\multicolumn{1}{|l|}{SQuAD v2.0} & \multicolumn{1}{l|}{Question Answering/Reading Comprehension} &\multicolumn{1}{l|}{130k} & \multicolumn{1}{l|}{11.9k} & F1/EM \\ \hline
\end{tabular}
\end{table}

\subsection{Tuning Procedure for MoEBERT and \modelename}
\label{supp-sec:moebert-tuning}
Following \cite{Zuo2022moebert}, we followed the  3-step process as outlined in the MoEBERT codebase\footnote{\url{https://github.com/SimiaoZuo/MoEBERT}}:
\begin{itemize}
\item We finetuned BERT on each downstream task for a set of 50 random hyperparameter trials over the following set:
\begin{itemize}
    \item Learning Rate: Discrete uniform over the set $\{1\times10^{-5}, 2\times10^{-5},3\times10^{-5},5\times10^{-5}\}$
    \item Batch size: Discrete uniform over the set $\{8,16,32,64\}$
    \item Weight Decay: Discrete uniform over the set $\{0,0.01,0.1\}$
    \item Epochs: 10
\end{itemize}
Note that this step matched the performance numbers reported for BERT-base in Table 1 of \citep{Zuo2022moebert}. We used the best model (for each dataset) for the remaining steps below.
\item Compute importance weights in FFN layers to construct an MoEBERT/COMET-BERT model, where FFN layers are replaced with MoE layers with the weight assignment strategy in \citep{Zuo2022moebert}.
\item Distill BERT into MoEBERT or \modelename~ on the downstream task with a layer-wise discrepancy loss.
For MoEBERT, we used the optimal hyperparameters reported (based on $\sim1000$ trials per dataset) in Table 7 of Supplement in \citep{Zuo2022moebert}. 
For \modelename, we performed 100 tuning trials via random search with each \modelaname~and \modelbname~and picked the best results based on development datasets. The hyperparameters were randomly selected from the following sets:
\begin{itemize}
    \item Learning Rate: Discrete uniform over the set $\{1\times10^{-5}, 2\times10^{-5},3\times10^{-5},5\times10^{-5}\}$
    \item Batch size: Discrete uniform over the set $\{8,16,32,64\}$
    \item Weight Decay: Discrete uniform over the set $\{0,0.01,0.1\}$
    \item Distillation Regularization ($\lambda_{distill}$ in \citep{Zuo2022moebert}): Discrete uniform over the set $\{1,2,3,5\}$.
    \item $\gamma$ (for smooth-step for \modelaname): Discrete uniform over the set $\{0.01,0.1,1.0\}$.
    \item $\lambda$ (for entropy regularization for \modelaname): Discrete uniform over the set $\{0.05, 0.1, 0.5, 1, 5, 10\}$.    
    \item Epochs: 50 for small datasets (CoLA, RTE, MRPC) and 25 for large datasets (SST-2, MNLI, QQP, QNLI, SQuADv2.0). Best model was recovered on development set on best checkpoint.
\end{itemize}
\end{itemize}


\end{document}